\documentclass[letterpaper,10pt,twocolumn]{article}

\usepackage[letterpaper,top=0.875in,bottom=0.875in,left=0.75in,
            right=0.75in,columnsep=0.25in]{geometry}

\usepackage{mathptmx}
\usepackage[T1]{fontenc}
\usepackage[utf8]{inputenc}

\usepackage[hyphens]{url}
\usepackage{graphicx}
\usepackage{natbib}
\usepackage{caption}
\usepackage{algorithm}
\usepackage{algorithmic}
\usepackage{booktabs}
\usepackage{amsmath}
\usepackage{amssymb}

\usepackage[compact]{titlesec}
\titleformat{\section}{\normalfont\large\bfseries}{\thesection}{0.6em}{}
\titleformat{\subsection}{\normalfont\normalsize\bfseries}{\thesubsection}{0.6em}{}
\titlespacing*{\section}{0pt}{1.6ex plus 0.6ex minus 0.2ex}{0.9ex plus 0.2ex}
\titlespacing*{\subsection}{0pt}{1.3ex plus 0.5ex minus 0.2ex}{0.7ex plus 0.2ex}

\title{\vspace{-1.2em}%
  \includegraphics[height=21mm]{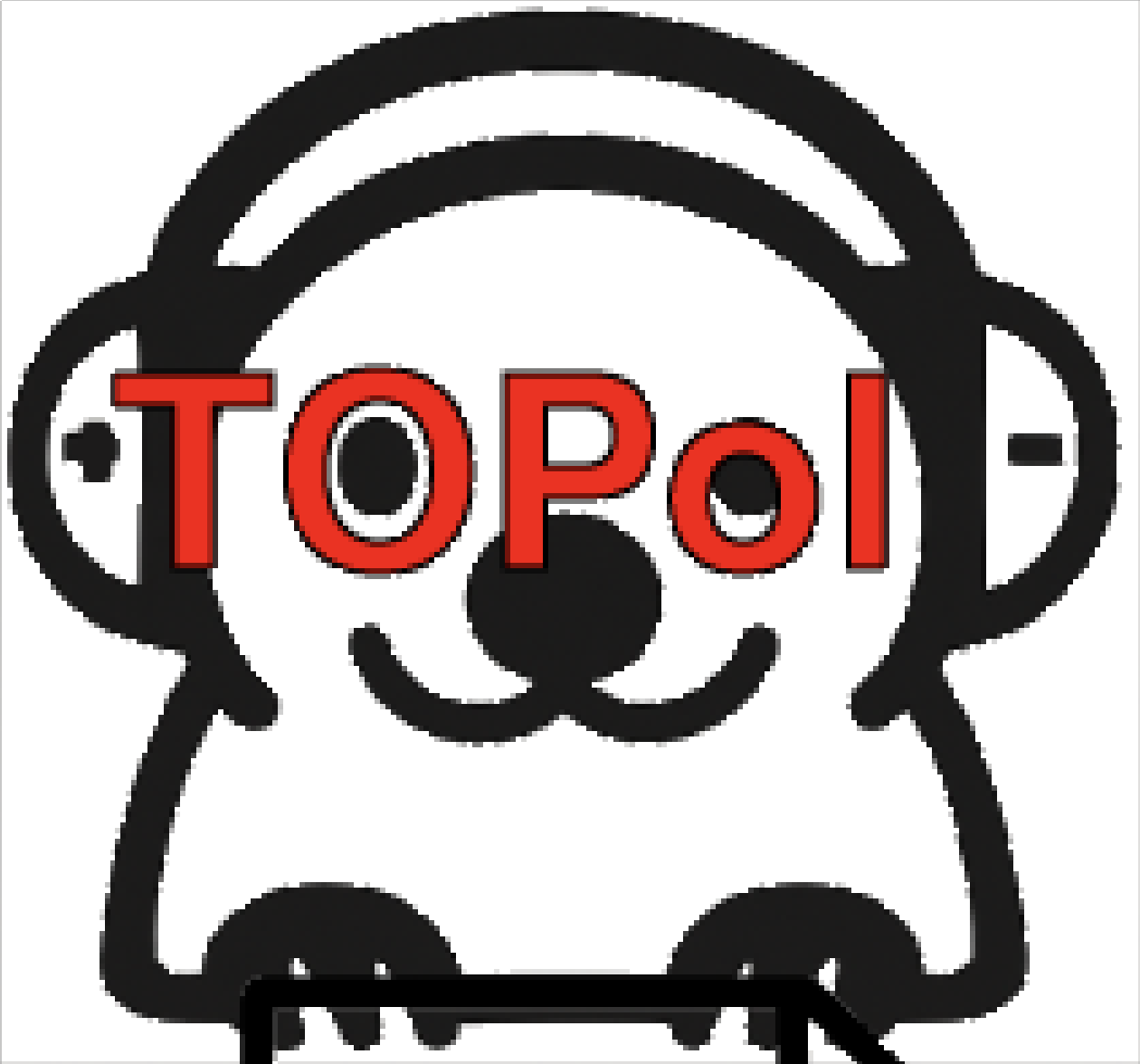}\\[2pt]
  TOPol: Reconstructing and Interpreting\\Multidimensional Semantic Polarity Fields}

\author{%
  Lucia Gomez\textsuperscript{1,2} \quad
  Gabin Taibi\textsuperscript{1,3} \quad
  Julius Sipke Kooistra\textsuperscript{1}\\[5pt]
  \normalsize\textsuperscript{1}Institute of Data Science and Finance,
  Bern University of Applied Sciences, Switzerland\\
  \normalsize\textsuperscript{2}Geneva Finance Research Institute,
  University of Geneva, Switzerland\\
  \normalsize\textsuperscript{3}Faculty of Behavioral Sciences,
  University of Twente, Netherlands\\[4pt]
  \normalsize Code and data: \url{https://github.com/gomez-L/TOPol}}

\date{}

\begin{document}

\maketitle

\begin{abstract}
Polarity in Natural Language Processing has long been operationalized as a single axis: the sentiment construct. We challenge this one-dimensional view -- not by proposing a second, better-named construct, but by asking what a topic's documents are bipolar \emph{about}, without specifying an answer in advance. TOPol inverts the usual concept-direction paradigm -- rather than mapping a \emph{named} construct to one direction, it maps a \emph{topic} to the full set of statistically reproducible bipolar semantic axes its documents support in a transformer's residual stream, and asks only afterwards which of them, if any, correspond to a known construct. These axes need not be evaluative: two poles can contrast subject matter, register, or which of two groups a review is about, just as easily as good versus bad. Method: a recursive, orthogonal-by-construction bipartition of each topic's document cloud, with every candidate axis promoted to a finding only if it survives a single pre-specified split-half stability gate. We calibrate discovery against a known answer before trusting it: on $9{,}986$ Amazon reviews ($14$ discovered topics) in Gemma 3 4B, we use the star-rating diff-in-means vector as a \emph{ruler} -- a direction known by construction to be sentiment -- and measure how much of the discovered geometry it explains. Discovery blindly re-derives the sentiment axis in \emph{every} topic (cosine $0.71$--$0.97$) without ever seeing the rating label, while the same stability gate certifies $543$ axes overall, of which $106$ are near-orthogonal to sentiment yet equally reproducible; per topic, $33$--$45$ mutually orthogonal axes survive, and sentiment axes resolve in later layers than the rest. What the non-sentiment axes actually contrast is tested three independent ways -- differential vocabulary, an internals-grounded natural-language decoder, and a vocabulary-based similarity clustering across all $543$ axes -- and, on a hand-checked sample, the three agree: e.g.\ one axis cleanly separates computing-peripheral reviews from audio-equipment reviews within a single electronics topic, confirmed by its distinctive words \emph{and} by which real documents its learned embedding independently places nearest each pole. Clustering axes by this shared vocabulary (not by raw direction) surfaces $99$ families, $9$ of them recurring across topics with human-readable labels -- a direct, checkable answer to which semantic contrasts are candidate corpus-wide regularities and which are topic-local, replacing a geometric proxy (raw-direction clustering pooled at one fixed layer) that we show conflates axes with genuinely different content. A further test -- correlating each axis's direction-similarity to the sentiment ruler against an off-the-shelf sentiment classifier that never touches the discovered geometry -- confirms direction alone is a weak, topic-heterogeneous proxy for meaning ($r=0.243$, $p{=}9\times10^{-9}$ pooled; $r=0.69$ to $-0.36$ across topics), reinforcing that most of this structure is not sentiment wearing a different hat. Together these results support treating polarity as a property of a topic's representational geometry -- with a measurable rank, a per-axis depth, and an empirically checkable set of recurring versus topic-local semantic contrasts -- rather than as a single universal scalar.
\end{abstract}

\section{Introduction}

Polarity analysis in Natural Language Processing is still effectively synonymous with sentiment analysis. From Osgood's evaluative factor \citep{osgood1957measurement} through modern sentiment lexica \citep{turney2002thumbs,hutto2014vader,pang2008opinion} to concept-direction and representation-engineering work that recovers a single vector for a single named construct \citep{mikolov2013distributed,bolukbasi2016man,kim2018interpretability,zou2023representation,turner2023steering,rimsky2024steering}, the field has treated ``polarity'' as one axis: good versus bad. This equation of polarity with sentiment is a historical convention -- the sentiment axis was named before it was measured -- and it carries a blind spot by design: any bipolar distinction nobody thought to name in advance is invisible to a method that must be handed a label before it can find a direction. Crucially, the missing distinctions need not be \emph{evaluative} at all. A topic's documents can be bipolar along an axis that has nothing to do with quality or sentiment -- reviews of a religious text written from a Christian versus a secular standpoint, a product discussed for infant versus adult use, a complaint about safety versus one about price -- and a method built to find ``the'' evaluative axis has no way to notice any of this. TOPol is built to find \emph{whatever} bipolar structure a topic's documents support, evaluative or not, and only afterwards ask which of it lines up with a known construct like sentiment.

TOPol inverts this concept-direction paradigm. Existing methods map a label to a direction: supply a concept, recover its vector. TOPol maps a discourse region to a \emph{set} of directions: supply a topic, recover every axis along which its documents are reproducibly bipolar, and ask only afterwards which construct each axis corresponds to. Under this inversion, semantic polarity becomes a property of a topic's representational geometry rather than of a lexicon, and two quantities that a single supervised vector cannot even express become measurable: the \emph{rank} of a topic's polarity field (how many independent bipolar axes its documents support) and, for each axis, the \emph{depth} in the network at which it is most sharply expressed.

The methodological difficulty with any such discovery procedure is trust: an unsupervised method that finds many directions in a high-dimensional cloud will always find \emph{something}, so the burden is to show that what it finds is reproducible and that, when the answer is known, it recovers it. We therefore evaluate TOPol first in a \emph{calibration} setting rather than a discovery setting. We run it on a sentiment-saturated corpus, Amazon reviews, where the primary axis of contrast is known by construction: the star rating. Treating the diff-in-means vector between low- and high-rated reviews as a \emph{ruler} -- a direction we know is sentiment -- we ask how much of the discovered geometry the ruler explains, and how much it misses, and -- for whatever it misses -- what it actually is. This is deliberately the case most favourable to the sentiment-is-polarity equation: if even here discovery surfaces stable, nameable structure orthogonal to the ruler, the one-axis view is incomplete on its own home ground. The calibration logic is the point; applying TOPol where the primary axis is \emph{unknown} is the companion discovery study this design is ultimately for (Section~\ref{sec:future}).

Two further claims are tested directly. (a) Each axis has a \emph{depth of maximal expression} -- a transformer layer at which its bipolar separation is sharpest -- so representational layer is a property measured per axis rather than a hyperparameter fixed in advance; whether that depth also indexes level of abstraction is a separate, interpretive question we raise but do not causally settle. (b) Axes that recur across topics, in the specific vocabulary they contrast rather than merely in raw direction, are candidates for corpus-wide semantic regularities, while topic-local ones are domain-specific constructs -- turning a geometric procedure into the beginning of an empirical typology of the semantic contrasts a corpus contains, of which sentiment is one member, not the definition of the category.

\paragraph{Contributions.}
(1) \textbf{A formal object.} We define the \emph{semantic polarity field} of a topic as the set of reproducible bipolar axes its document manifold supports, each carrying a direction, a depth profile, and a stability certificate; against this object, per-topic \emph{rank} is an exact count (guaranteed by within-topic deflation) and cross-topic \emph{dimensionality} is a participation-ratio estimate. This gives polarity a measurable rank and depth spectrum where prior work had a single scalar.
(2) \textbf{A discovery procedure with a trust guarantee}: a recursive, orthogonal-by-construction bipartition over a real transformer's residual stream in which nothing is reported unless it survives one pre-specified split-half stability gate, whose selectivity we argue from the near-zero cosine of random directions in a $2560$-dimensional space.
(3) \textbf{A calibration result}: discovery blindly recovers the known sentiment construct in all $14$ topics while, in the same pass, certifying $106$ near-orthogonal axes as equally reproducible -- and, surprisingly, sentiment is the top-variance direction in only $1$ topic.
(4) \textbf{A structural account}: where this excess structure lives (a depth separation between sentiment and non-sentiment axes), and, via a vocabulary-based similarity clustering over all $543$ axes (not raw direction), which specific semantic contrasts recur across topics ($9$ families, with human-readable labels) versus which are topic-local ($90$ of $99$ families) -- a result we show a purely geometric (raw-direction, single-layer) clustering cannot deliver, because it groups axes that share almost no vocabulary and separates axes that share a great deal.
(5) \textbf{A three-way interpretation cross-check}: a per-axis, per-topic measurement of whether direction-similarity to the ruler predicts an \emph{independent} sentiment estimate; differential-vocabulary and language-model readings of the excess axes; and an internals-grounded natural-language decoder (trained to reconstruct each pole's own text from its activation) whose learned-embedding nearest neighbours corroborate the vocabulary reading on hand-checked axes. Throughout, we position TOPol against the two nearest paradigms -- supervised concept directions, which name before measuring, and sparse autoencoders, which learn a \emph{global} feature dictionary rather than a \emph{topic-local, bipolar, depth-resolved} axis set (Section~\ref{sec:related}).

\section{Related Work}
\label{sec:related}

\paragraph{Sentiment Polarity as one axis.} Osgood's semantic differential identified evaluation as the dominant factor in how people rate concepts \citep{osgood1957measurement}, and the sentiment-analysis literature that followed operationalized that single factor at scale, from lexicon-based scoring \citep{turney2002thumbs,hutto2014vader} to supervised and rating-derived corpora \citep{mcauley2013hidden,zhang2015characterlevel} and comprehensive surveys \citep{pang2008opinion}. Reviews, in particular, are almost always reduced to a sentiment-rich evaluative scalar (a star rating or a lexicon score), while TOPol tests whether that scalar is unsupervisely represented in reviews, and how it relates with possibly many non-sentiment related polar axes.

\paragraph{Concept directions and representation engineering.} A parallel line of work shows that a wide range of human-nameable concepts correspond to linear directions in a network's representation space, recovered by supplying the concept as a label: word-analogy directions \citep{mikolov2013distributed}, a gender-bias direction found by contrasting labelled word pairs \citep{bolukbasi2016man}, concept activation vectors fit from user-supplied examples \citep{kim2018interpretability}, and, more recently, activation-addition and representation-engineering methods that read off and add back a labelled contrast's direction to steer generation \citep{turner2023steering,rimsky2024steering,zou2023representation}. Every one of these methods maps a \emph{given} label to a direction. TOPol runs the inverse operation: it does not start from a label, but instead asks what directions a topic's own documents support, aiming at labeling only afterwards.

\paragraph{Unsupervised direction discovery and the granularity question.} The paradigm closest to TOPol is unsupervised, not supervised: methods that read structure out of the representation space without being handed a label. The linear representation hypothesis holds that many concepts are encoded as linear directions \citep{mikolov2013distributed,park2024linear}, with superposition packing more features than dimensions \citep{elhage2022superposition}; sparse probing \citep{gurnee2023finding} and, most prominently, sparse autoencoders (SAEs) \citep{cunningham2023sparse,bricken2023monosemanticity,templeton2024scaling} recover large dictionaries of monosemantic features. A natural reviewer question is why not simply run an SAE, or per-topic PCA, and read off the directions. The answer is that these methods answer a different question at a different granularity. SAEs learn a single, corpus-\emph{global} dictionary of \emph{unipolar} features (directions of \emph{presence}) shared across the whole model; PCA returns a topic's top-\emph{variance} directions -- which, as our depth-1 result shows (Section~\ref{sec:results-id}), are usually \emph{not} the sentiment axis at all, let alone whatever other bipolar structure a topic supports. TOPol instead recovers, per topic, the set of \emph{bipolar contrast} axes (directions of opposition, not presence), each gated by an explicit reproducibility test and resolved to a depth. The unit of interpretation -- topic-local, bipolar, stability-gated, depth-resolved -- sits between the supervised single concept vector and the global SAE feature, and is not produced by either. The two are complementary rather than competing: decomposing a discovered TOPol axis into SAE features is a natural further interpretation route (Section~\ref{sec:future}), not an alternative to the internals-grounded route we do deploy here (next paragraph).

\paragraph{Mechanistic interpretability and layer-wise structure.} Layer-wise probing has shown that different depths of a network encode systematically different levels of linguistic structure \citep{tenney2019bert}; TOPol's depth profile extends this from a single, labelled probe to every discovered axis, treating layer as a quantity measured per direction rather than fixed in advance. Beyond the document-level readings this affords, we additionally train a small internals-grounded natural-language decoder (Section~\ref{sec:method-part2}) that reconstructs a pole's own text from its activation through a compact latent, and evaluate a separate head's fidelity at reconstructing the activation itself back out -- an interpretation signal read from the residual stream, not only from which documents sit at an axis's extremes. Full sparse-autoencoder decomposition against a released dictionary for the same model is a further, deeper interpretation route we do not deploy in this calibration study; we return to it in Section~\ref{sec:future}.

\section{Method}
\label{sec:method}

\paragraph{Overview.} The pipeline has two estimators that must be kept conceptually distinct. The first, the \emph{ruler}, is a single supervised direction computed exactly as prior diff-in-means / concept-direction work would compute it (Section~\ref{sec:method-cb}); its semantics are known by construction to be sentiment. The second, the \emph{object being measured}, is TOPol's discovery core: with no access to any label it recovers every reproducible bipolar direction a topic's documents support (Sections~\ref{sec:method-discovery}--\ref{sec:method-gate}). The ruler is never used to \emph{find} anything; it is used only \emph{afterwards} to ask which discovered axis it matches (identification) and how much stable structure lies away from it (excess). Each subsection below is tagged as belonging to the ruler or to the object, and every downstream analysis (Sections~\ref{sec:method-field}--\ref{sec:method-part2}) operates on the object with the ruler held out.

\paragraph{Problem formulation.} Fix a topic $t$ with document activations cached at a set of layers $\mathcal{L}$. We define the \textbf{semantic polarity field} of $t$ as a set of axes
\begin{equation}
\mathcal{F}_t = \{\, (\hat u_j,\; d_j(\cdot),\; \ell^\ast_j,\; s_j) \,\}_{j=1}^{r_t},
\end{equation}
where each axis carries a unit direction $\hat u_j$, a depth profile $d_j(\ell)$ giving its bipolar effect size at each layer $\ell\in\mathcal{L}$, a peak layer $\ell^\ast_j=\arg\max_\ell d_j(\ell)$, and a stability certificate $s_j\in\{0,1\}$ from the gate of Section~\ref{sec:method-gate}. Only axes with $s_j=1$ enter $\mathcal{F}_t$. Two dimensionality measures are defined against this object, and they measure deliberately different things. \textbf{Per-topic rank} $r_t=\lvert\mathcal{F}_t\rvert$ is an \emph{exact} count of mutually orthogonal bipolar axes, because within-topic discovery deflates each direction before seeking the next (Algorithm~\ref{alg:discovery}); orthogonality is a construction guarantee, not an empirical estimate. \textbf{Cross-topic dimensionality} pools axes from \emph{all} topics -- which are \emph{not} mutually deflated and therefore not orthogonal by construction -- and estimates the effective number of shared directions via a participation ratio (Section~\ref{sec:method-field}); it is an approximation precisely because the cross-topic ensemble carries no orthogonality guarantee. Keeping these two apart is what dissolves the apparent tension between high per-topic rank and low pooled dimensionality reported in Section~\ref{sec:results-families}: the first counts local structure, the second counts shared structure. The load-bearing theoretical element throughout is the stability gate: spherical $2$-means always returns \emph{a} bipartition, so a found split is not yet an axis; the gate is what promotes a reproducible split to a member of $\mathcal{F}_t$, and its selectivity (Section~\ref{sec:method-gate}) is what the field's meaning rests on.

\subsection{Model and Representation Space}

Every step that touches meaning runs on one open model, Gemma 3 4B, pretrained (base) variant, \texttt{google/gemma-3-4b-pt} \citep{gemma2025gemma3}, $d_{\text{model}}=2560$. Using one fixed, open model throughout is a design constraint: the model whose activations carry the field is also the model any future causal or decomposition-based interpretability (Section~\ref{sec:discussion}) would probe, so nothing downstream would be explaining a different system from the one measured. We use the pretrained, not instruction-tuned, variant because instruction tuning writes persona and chat-template structure into the same residual stream the polarity vectors are computed from, and because chat-template/special-token positions carry disproportionately large activation norms that would otherwise dominate a mean-pooled document vector. The residual stream is cached at every second layer plus the endpoints ($0,2,4,\dots,34$; layer $0$ is the embedding output), 18 checkpoints in total, as float16 mean-pooled document vectors, explicitly excluding BOS and special/chat-template token positions from the pool ($9{,}986$ of $962{,}467$ pooled-token positions, $1.04\%$, excluded corpus-wide).

\subsection{Corpus}

We use $9{,}986$ Amazon product reviews (of $10{,}000$ raw, after a $\geq 100$-character filter), with a boundary built from the star rating: regime A = ratings $\leq 2$ ($n=4{,}993$), regime B = ratings $\geq 4$ ($n=4{,}993$), 3-star reviews dropped. This corpus is chosen because its primary contrast is \emph{known}: if a blind discovery procedure cannot recover product-review sentiment, it has no business being trusted where the answer is unknown. The corpus preparation used here exposes a precomputed binary polarity label but not the underlying continuous star rating, so wherever we need a signal independent of the discovery procedure and of that label's own geometry, we substitute an off-the-shelf sentiment classifier never involved in discovery (Section~\ref{sec:method-part2}).

\subsection{Topic Segmentation}

Topics are assigned independently of the field model, by a dedicated sentence encoder (EmbeddingGemma-300m) feeding a $k$-nearest-neighbour graph ($k=15$, cosine metric, 5 components, built with UMAP \citep{mcinnes2018umap}) partitioned by the Leiden algorithm \citep{traag2019leiden} (resolution $1.0$). Decoupling topic identity from the field model avoids a circularity in which the partition depends on the very space being probed for polarity structure. Because fitting the partition on the pooled corpus lets the size of a topic's own regime-A/B displacement bias which topics survive (large-displacement topics get split into two communities by Leiden and silently dropped when one side is left unpopulated), segmentation is fit on regime-A documents only (the UMAP graph and Leiden partition are both computed from regime-A EmbeddingGemma embeddings alone), and each regime-B document is then assigned to the regime-A cluster whose centroid -- computed in that same EmbeddingGemma embedding space, not the Gemma field space -- it is closest to by cosine similarity. This yields 15 communities; the smallest (15 documents, entirely regime B) is dropped before any regime-comparison step, leaving the $\mathbf{14}$ topics analysed throughout this paper ($220$--$1{,}285$ documents each, $9{,}971$ documents total).

\subsection{[Ruler] The Contextual-Boundary Estimator}
\label{sec:method-cb}

\begin{figure}[t]
\centering
\includegraphics[width=0.95\columnwidth]{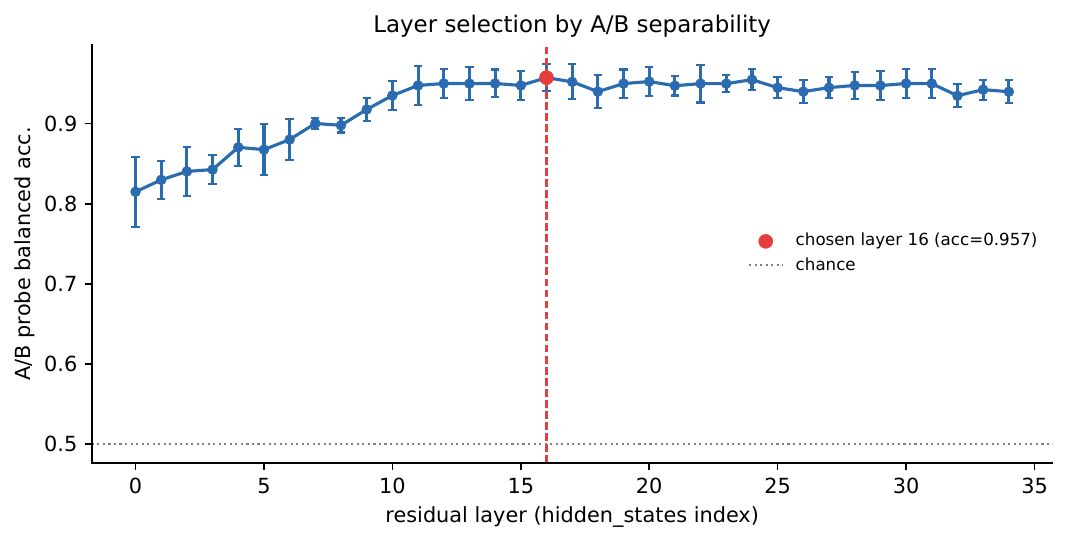}
\caption{Regime (A/B) probe balanced accuracy by residual layer, on a 400-document subsample. The broad plateau from layer $\sim$10 onward means sentiment separability is not confined to a narrow depth band; layer 16 (chosen, red) sits inside it.}
\label{fig:layer-selection}
\end{figure}

For a topic $t$ with regime-A and regime-B document sets, the contextual-boundary (CB) vector is the difference of centroids,
\begin{equation}
v^t_{CB} = \bar\mu^t_B - \bar\mu^t_A ,
\end{equation}
computed in the full $2560$-dimensional representation space at one fixed reference layer, chosen by 5-fold cross-validated logistic-probe balanced accuracy separating regime A from regime B on a $400$-document subsample across all 35 layers (Figure~\ref{fig:layer-selection}): accuracy rises from $0.815$ at the embedding layer to a broad plateau of $0.93$--$0.96$ from roughly layer 10 onward. We select layer $\mathbf{16}$ (balanced accuracy $0.957$) as the reference layer $\ell_{\text{ref}}$ used for $v^t_{CB}$ and for all cross-topic pooling below. This is exactly the standard diff-in-means / concept-direction estimator \citep{mikolov2013distributed,bolukbasi2016man,turner2023steering}, computed once per topic; TOPol changes nothing about how $v^t_{CB}$ itself is built.

\subsection{[Object] Axis Discovery: Recursive Bipartition}
\label{sec:method-discovery}

Independent of $v^t_{CB}$ and of the regime label entirely, every cached layer of every topic's document cloud is searched for reproducible two-way structure (Algorithm~\ref{alg:discovery}). The two-way split is the natural primitive here because the object of interest is \emph{bipolarity}: an axis is by definition a direction of opposition between two poles, so a $2$-means contrast -- rather than a $k$-way clustering or a full PCA basis -- is the minimal unit that matches the construct, and spherical ($\ell_2$-normalised) $2$-means is chosen because direction, not magnitude, defines a polarity axis. At one topic, one layer: centre the documents on their own mean; split the centred cloud in two by spherical (cosine) 2-means; record the between-means direction and its effect size
\begin{equation}
d(\ell) = \frac{\lVert v(\ell)\rVert}{\sigma_{\text{pooled}}(\ell)}, \qquad
\sigma_{\text{pooled}}(\ell) = \sqrt{\tfrac{1}{n}\textstyle\sum_i \lVert x_i - \mu_{g(i)}\rVert^2},
\end{equation}
where $g(i)\in\{A,B\}$ is document $i$'s side of the split and the sum runs over both sides' within-split residuals. This $d(\ell)$ is a between-pole separation normalised by pooled within-pole scatter -- a Cohen's-$d$ / Fisher-ratio analog -- so it is directly comparable across axes and layers and is what the depth profile tracks. We then project the between-means direction out of the (still-centred) cloud and repeat up to $\texttt{max\_depth}=4$ times, stopping early if either side of a split falls below $\texttt{min\_cluster\_size}=30$ documents. Because each split is found only after every previous direction has been projected out, the directions recorded for one topic at one layer are \emph{orthogonal to one another by construction} -- this is what makes per-topic rank an exact count. Candidate splits found at adjacent layers that represent the same underlying partition (adjusted Rand index $\geq \texttt{ari\_merge}=0.7$ on the document labelings, average linkage over $1-\text{ARI}$ distance) are merged into one \emph{contrast}, each carrying a full depth profile $d(\ell)$ across all 18 cached layers and a peak layer $\ell^\ast = \arg\max_\ell d(\ell)$. The merge threshold $0.7$ is a fixed hyperparameter chosen to be conservative (requiring substantial label agreement before two layers' splits are treated as one axis); it trades axis count against redundancy and is reported with the run configuration (Appendix~\ref{app:config}) rather than tuned against any outcome.

\begin{algorithm}[tb]
\caption{Per-topic, per-layer axis discovery}
\label{alg:discovery}
\textbf{Input}: topic documents' raw activations $X_\ell \in \mathbb{R}^{n\times d}$ at layer $\ell$\\
\textbf{Parameters}: \texttt{max\_depth}, \texttt{min\_cluster\_size}\\
\textbf{Output}: list of candidate contrasts, each with a direction and effect size
\begin{algorithmic}[1]
\STATE $X_r \gets X_\ell - \text{mean}(X_\ell)$
\FOR{$\text{depth} = 1$ to \texttt{max\_depth}}
  \STATE $(\text{labels}) \gets$ spherical 2-means on $X_r$
  \IF{$\min(n_A, n_B) < \texttt{min\_cluster\_size}$}
    \STATE \textbf{break}
  \ENDIF
  \STATE $v \gets \text{mean}(X_\ell[\text{labels}{=}B]) - \text{mean}(X_\ell[\text{labels}{=}A])$
  \STATE $d \gets \lVert v \rVert / \sigma_{\text{pooled}}$; record (labels, $v$, $d$, depth)
  \STATE $\hat u \gets v / \lVert v \rVert$
  \STATE $X_r \gets X_r - (X_r \hat u)\hat u^\top$ \COMMENT{project out $\hat u$, recurse}
\ENDFOR
\end{algorithmic}
\end{algorithm}

\paragraph{From per-layer candidates to one contrast with a depth profile.} Two distinct loops are involved here, and keeping them separate matters for what follows. \emph{First}, Algorithm~\ref{alg:discovery} is run \textbf{once per cached layer} -- 18 independent runs per topic, one for each $\ell\in\mathcal{L}$ -- each producing its own short sequence of up to \texttt{max\_depth} candidates at that layer alone, with no knowledge of what any other layer found. \emph{Second}, because the same underlying bipartition can resurface at several adjacent layers (a split that best separates the cloud at layer 14 is often still a good split at layers 16 and 18), candidates whose document labelings agree closely across layers (adjusted Rand index $\geq \texttt{ari\_merge}=0.7$, average linkage over $1-\text{ARI}$ distance) are merged into one \emph{contrast}; the member with the single highest effect size becomes that contrast's representative direction, and \texttt{layers\_found} records every layer that contributed to the merge. This \texttt{ari\_merge} threshold governs only this within-topic, cross-\emph{layer} deduplication of candidates -- it is a different question, scored on a different population of vectors, from the cross-\emph{half} reproducibility test in Section~\ref{sec:method-gate} (whose own, numerically different, threshold is $\texttt{stability\_threshold}=0.6$); the two should not be conflated even though both are ARI/cosine-style similarity cuts. Once a contrast's representative direction and document labels are fixed this way, its \textbf{depth profile} $d(\ell)$ is obtained by re-evaluating that one, fixed A/B partition's effect size at \emph{every} one of the 18 cached layers in turn -- i.e.\ the same set of documents stays assigned to pole A or pole B, and only the activations they are measured against change layer to layer. The depth profile is therefore not ``one $d$ value per layer where a new split happened to be found''; it is one already-discovered split's separation strength traced across the whole network, and $\ell^\ast=\arg\max_\ell d(\ell)$ is that curve's peak.

Run this way and merged within each topic, discovery yields a flat list of candidate contrasts per topic (mean $51.4$ per topic, range $47$--$57$; $719$ total across 14 topics), each carrying a discovered layer (where its representative candidate was found), a recursion depth, a full 18-point depth profile, a peak layer $\ell^\ast$, and an effect size at $\ell^\ast$. None of this is yet a finding: as the problem formulation states, a bipartition is not yet an axis of $\mathcal{F}_t$ until it survives the one filter this pipeline applies -- described next, and it is this filter's own output, not the count above, that Table~\ref{tab:topics} reports as ``Candidates $\to$ Passed.''

\subsection{[Object] The Split-Half Stability Gate}
\label{sec:method-gate}

Exactly one filter is applied to everything discovery proposes (Algorithm~\ref{alg:gate}). It is worth walking through in full, because every ``Candidates $\to$ Passed'' count reported in this paper (e.g.\ Table~\ref{tab:topics}'s $53\to45$ for Topic 0) is exactly this procedure's output, and the two numbers either side of the arrow answer different questions: the left number is how many bipartitions Section~\ref{sec:method-discovery} found at all; the right is how many of those the test below certifies as reproducible.

\emph{Three independent discovery runs, not one.} For a given topic, the full discover-and-merge procedure of Section~\ref{sec:method-discovery} (run at every cached layer, merged across layers) is executed \textbf{three separate times}: once on the topic's complete document set, giving $C_{\text{full}}$; once on a random $\sim\!50\%$ half $H_1$, giving $C_1$; and once, entirely independently, on the disjoint other half $H_2$, giving $C_2$. $H_1$ and $H_2$ are fixed once per topic (the same two halves for every layer and every recursion depth), and $C_1$/$C_2$ are not subsets or filtered versions of $C_{\text{full}}$ -- each is the result of re-running the whole discovery pipeline from scratch on less data, so the three runs can and do disagree: a topic's $C_1$ and $C_2$ typically differ from each other, and both from $C_{\text{full}}$, in how many candidates they find, in what order, and at which layers they peak.

\begin{figure}[t]
\centering
\includegraphics[width=0.95\columnwidth]{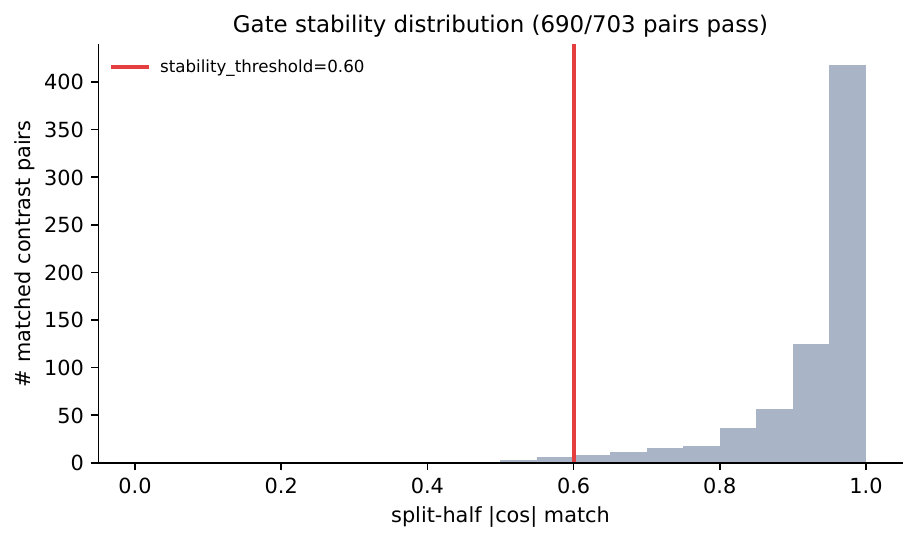}
\caption{Split-half $\lvert\cos\rvert$ for every greedily matched half-1/half-2 contrast pair, pooled across all 14 topics, against the stability threshold ($0.6$).}
\label{fig:gate-stability}
\end{figure}

\emph{Greedy matching.} Every contrast in $C_1$ is scored against every contrast in $C_2$ by $\lvert\cos\rvert$, taken at whichever of the pair's own two peak layers $\ell^\ast$ gives the larger value (so a real match is not penalised merely because the two halves disagree slightly on where it peaks). These pairwise scores are then resolved into a one-to-one assignment \emph{greedily}: the single highest-scoring $(c_1,c_2)$ pair across the whole score matrix is matched first and both contrasts removed from further consideration; the highest-scoring pair among what remains is matched next; and so on until one side is exhausted. This produces exactly one matched partner (or none) for every contrast in the smaller of $C_1$, $C_2$, and the pairwise score is computed and logged for \emph{every} resulting pair, not only the ones that go on to pass -- pooling all $703$ such pairs across the 14 topics is what Figure~\ref{fig:gate-stability} plots.

\emph{The pass/fail test.} A matched pair $(c_1,c_2)$ is \textbf{stable} only if $\lvert\cos(c_1,c_2)\rvert \geq \texttt{stability\_threshold}=0.6$: the same direction had to be rediscovered from scratch on two halves of the data that never saw each other's documents. This is the one and only filter in the entire pipeline -- nothing about a candidate's effect size, its layer, or its recursion depth is ever used to decide pass/fail, only this cross-half reproducibility score.

\emph{Mapping a stable pair back onto the full-topic contrast.} A stable pair does not itself become the reported axis. Instead, its $H_1$ member $c_1$ is matched onto whichever contrast in $C_{\text{full}}$ is closest to it by $\lvert\cos\rvert$ at $c_1$'s own $\ell^\ast$, and \emph{that} full-topic contrast -- not $c_1$, not $c_2$ -- is the one flagged \textbf{passed}. This indirection is deliberate: $C_{\text{full}}$ is estimated from twice as much data as either half, so its direction, effect size, and depth profile are better-powered than either half's own version of the same split; the half-vs-half match is used purely as a reproducibility \emph{certificate} that a given full-topic contrast is not a fluke, and the certificate, once earned, is attached to the full-topic estimate rather than replacing it. Every $C_{\text{full}}$ contrast that no stable pair maps onto keeps its candidate status but is flagged not-passed rather than deleted, which is why every output table in this paper -- not just the ones we quote -- carries both passed and not-passed rows, distinguishable by this one flag, so the threshold can be revisited against the full logged distribution rather than trusted blindly.

The threshold $0.6$ is far more stringent than it looks, and the argument is the geometry of high dimensions. Two \emph{independent} directions in the $d_{\text{model}}=2560$-dimensional representation space have expected absolute cosine on the order of $1/\sqrt{d}\approx 0.02$, with the probability of $\lvert\cos\rvert\ge 0.6$ arising by chance between random directions astronomically small. A gate at $0.6$ therefore does not sit at an arbitrary point on a uniform scale; it demands that a direction independently rediscovered on a disjoint half of the data reappear at a cosine roughly $30\times$ above the chance floor. The empirical distribution (Figure~\ref{fig:gate-stability}) confirms the gate operates in a sparse region well below the main mass of true matches near $1.0$, so it is simultaneously permissive of genuine axes and effectively impossible for noise to clear. (A direct selectivity baseline -- how many random or label-shuffled directions clear the gate -- is the natural confirmatory experiment and is flagged as future work in Section~\ref{sec:future}.)

\begin{algorithm}[tb]
\caption{Split-half stability gate (one topic)}
\label{alg:gate}
\textbf{Input}: topic's cached activations at all layers; \texttt{stability\_threshold}\\
\textbf{Output}: full-topic contrasts, each flagged passed/not-passed
\begin{algorithmic}[1]
\STATE $C_{\text{full}} \gets$ Algorithm~\ref{alg:discovery} on the full topic, every layer, merged
\STATE split documents into disjoint halves $H_1, H_2$
\STATE $C_1 \gets$ Algorithm~\ref{alg:discovery} on $H_1$; \ $C_2 \gets$ Algorithm~\ref{alg:discovery} on $H_2$
\STATE $M \gets$ greedy $\lvert\cos\rvert$-matching of $C_1$ against $C_2$ (each pair scored at the better of the two sides' own $\ell^\ast$)
\FOR{each matched pair $(c_1,c_2) \in M$ with $\lvert\cos(c_1,c_2)\rvert \geq \texttt{stability\_threshold}$}
  \STATE map $c_1$ onto its closest contrast in $C_{\text{full}}$ by $\lvert\cos\rvert$ at $c_1$'s own $\ell^\ast$; flag that full-topic contrast \textbf{passed}
\ENDFOR
\STATE all other members of $C_{\text{full}}$ remain flagged not-passed (retained, not deleted)
\end{algorithmic}
\end{algorithm}

Pooling every half-vs-half matched pair across all 14 topics, $690$ of $703$ pairs ($98.2\%$) clear the threshold, and the distribution is heavily concentrated near $1.0$ (Figure~\ref{fig:gate-stability}): the threshold sits in a sparsely populated region well below the main mass of matches, not at a knife-edge cut through a uniform distribution. Of the $719$ candidate contrasts, $\mathbf{543}$ ($75.5\%$) pass the gate; among these, split-half $\lvert\cos\rvert$ has mean $0.950$ (median $0.976$, min $0.630$) and effect size at $\ell^\ast$ has mean $0.826$ (median $0.683$, range $0.263$--$2.340$).

\subsection{[Object] Cross-Topic Field Assembly and Dimensionality}
\label{sec:method-field}

As set out in the problem formulation, two distinct notions of dimensionality are used, and they answer different questions by design: per-topic rank measures how much bipolar structure exists \emph{within} a topic, while cross-topic dimensionality measures how much of that structure is \emph{shared} across topics. A field that is rich within every topic but shares only a little structure across topics will therefore -- correctly, not paradoxically -- show high per-topic rank and low cross-topic (pooled, single-layer) dimensionality simultaneously; Section~\ref{sec:results-families} finds exactly this pattern, together with a direct, vocabulary-based account of what little structure IS shared, and a caution about reading too much into the pooled-geometry number itself. \textbf{Per-topic rank} $r_t$ is simply the number of gate-passing contrasts in topic $t$; because within-topic discovery enforces orthogonality by construction (Algorithm~\ref{alg:discovery}, line 9), this is an exact rank count, not an approximation. \textbf{Cross-topic dimensionality} pools every gate-passing axis from every topic, evaluated at the single fixed reference layer $\ell_{\text{ref}}=16$, into one $k\times d$ ensemble $V$ and summarises its uncentred second-moment spectrum with the participation ratio,
\begin{equation}
\mathrm{PR}(V) = \frac{\left(\sum_i \lambda_i\right)^2}{\sum_i \lambda_i^2} \in [1,k], \qquad \lambda_i = \sigma_i(V)^2 ,
\end{equation}
the effective number of independent directions in the pooled ensemble. $\mathrm{PR}$ is bounded and directly interpretable at both ends: $\mathrm{PR}=1$ means every one of the $k$ pooled axes points (up to sign) the same way, and $\mathrm{PR}=k$ means all $k$ are mutually orthogonal; a value close to either bound licenses the corresponding reading directly, and it is only intermediate values that require care. The spectrum is deliberately \emph{uncentred}: a field in which every topic's axis pointed the same way is the canonical one-dimensional case, and mean-centring across topics would erase exactly that shared direction rather than reveal it. Because axes from different topics are not constructed orthogonal to one another (only within-topic deflation guarantees that), the pooled ensemble is additionally clustered into \textbf{families} by $\lvert\cos\rvert$ similarity of the axis vectors (average linkage on $1-\lvert\cos\rvert$ distance, merge threshold $0.7$ -- the same value, and the same conservative-merge rationale, as \texttt{ari\_merge} in Section~\ref{sec:method-discovery}, reused here for consistency of interpretation even though it is applied to a different population of vectors: \texttt{ari\_merge} deduplicates one topic's own candidates across cached \emph{layers}, whereas this step groups already gate-passed axes across all 14 \emph{topics} by raw direction). This is a purely geometric notion of family -- it groups axes that point the same way in the residual stream, and says nothing by itself about what either pole actually contains. Section~\ref{sec:method-part2} introduces a second, vocabulary-based family clustering that asks a meaning-based version of the same question directly from each axis's own text; Section~\ref{sec:results-families} reports both rather than treating one as a refinement of the other, because, as that section shows, the two clusterings largely disagree.

\subsection{[Ruler $\times$ Object] Identification and Excess Structure}
\label{sec:method-compare}

For every topic, each gate-passing axis is compared to $v^t_{CB}$ by $\lvert\cos\rvert$ at $\ell_{\text{ref}}$. The best-matching axis -- i.e.\ $\arg\max$ over that topic's gate-passing axes of $\lvert\cos\rvert$ to $v^t_{CB}$ -- is the \textbf{identification} axis, exactly one per topic by construction: evidence, if the match is strong, that discovery blindly re-derived the known construct. Every other gate-passing axis with $\lvert\cos\rvert < \texttt{excess\_cos\_threshold}=0.3$ to $v^t_{CB}$ is counted as \textbf{excess structure}: statistically stable contrastive structure that is not simply a noisy copy of the ruler. The threshold $0.3$ is deliberately strict against false excess: it excludes not only near-parallel axes but any axis retaining moderate alignment to the ruler, so an axis counted as excess shares little more than chance direction with sentiment (recall the chance floor is $\approx 0.02$ in this space); an axis at $\lvert\cos\rvert=0.3$ still shares under $10\%$ of its squared projection with the ruler. Excess is thus a conservative label for ``genuinely not the sentiment direction,'' not a lenient one.

These two labels are not exhaustive, and the arithmetic in Table~\ref{tab:topics} makes the gap explicit: of a topic's $n_{\text{passing}}$ gate-passing axes, exactly $1$ is identification and \texttt{excess\_count} are excess, but the remaining $n_{\text{passing}}-1-\texttt{excess\_count}$ -- typically the majority; e.g.\ $45-1-8=36$ for Topic 0 -- sit at intermediate $\lvert\cos\rvert$ to the ruler. They are neither singled out as ``the'' sentiment axis nor certified as orthogonal to it, and this section says nothing about what they individually mean; Section~\ref{sec:results-qual} reads a sample of them (drawn without regard to this cosine bucketing) directly from their text instead of relying on where they happen to fall on this one scale.

\subsection{Interrogating the Discovered Geometry}
\label{sec:method-part2}

Everything above discovers and validates directions without saying what any of them \emph{mean}. We use four complementary, non-causal probes of the discovered geometry, all of which stop short of decomposing what network features compose a direction or of testing what happens under intervention (Section~\ref{sec:discussion} discusses that next step). The first three read a single axis in isolation; the fourth asks whether two axes -- possibly from different topics -- mean the same thing.

\paragraph{Differential vocabulary.} For each topic, a single TF-IDF vectorizer (unigrams and bigrams, English stop-words removed, $\texttt{min\_df}=2$, $3{,}000$ features) is fit once on that topic's own documents and reused for every gate-passing axis in that topic, each time re-split by that axis's own pole labels -- valid because every axis of a topic visits the same document order, only the partition differs. Terms are ranked by the difference of mean TF-IDF between an axis's two poles, so a term is reported only on the side it is actually more characteristic of; the top 20 terms per pole are kept per axis, for all $543$ axes. Of the four probes, this is the only one that involves no model generation at all, and we treat it as the most reliable single signal for that reason.

\paragraph{Grounded language-model glosses.} For each axis, four real excerpts per pole (truncated to 40 words) plus that axis's own top TF-IDF words are assembled into a completion-style prompt -- \emph{not} an instruction -- ending mid-sentence, and completed by the same \texttt{gemma-3-4b-pt} checkpoint already used for the field (max 50 new tokens, cleaned to its first two sentences). A completion prompt is used because the pretrained checkpoint is not instruction-tuned and reliably continues text rather than obeys a directive; grounding the prompt in real excerpts keeps the model's output tied to the corpus rather than free-associated. We also built the identical grounded-evidence prompt as a direct instruction to \texttt{gemma-3-4b-it}, the instruction-tuned sibling checkpoint, as a planned comparison route (Section~\ref{sec:future}); it is not run for the results reported here.

\paragraph{Internals-grounded decoding (NLA/AR).} A small VAE-style natural-language autoencoder (NLA) -- a two-layer MLP encoder mapping an activation to a 256-dimensional latent, and a GRU decoder mapping that latent back to a token sequence -- is trained to reconstruct each document's own (truncated, whitespace-tokenised) text from its activation \emph{at that axis's own discovered layer $\ell^\ast$}. There is exactly \textbf{one} encoder/decoder pair for the entire run, not one per axis or per topic: every (axis, document) pair from every one of the 543 axes is pooled into a single training set, and the encoder/decoder are conditioned on a per-row axis-id and layer-id embedding (concatenated into the network's input/hidden state) so one shared model can still specialise its reconstruction to each axis's own layer and vocabulary, rather than 543 separately-trained small models that could not share statistical strength across axes (cross-entropy reconstruction plus a KL term toward a standard-normal prior). A separately trained regression head (AR) is evaluated, under 5-fold cross-validation, on how faithfully the frozen encoder's 256-dimensional bottleneck lets the original activation be reconstructed. Two outputs of this route are used here: (i) each pole's \emph{centroid} -- the mean of the encoder's latent embedding over every document assigned to that pole, by that axis's own discovered A/B partition, not by the corpus's regime label -- decoded into a prototype description \emph{and} grounded against its 3 nearest real documents in that same latent space (nearest by Euclidean distance between the document's own encoded latent and the pole centroid; these real documents, not the decoder's free generation from the centroid itself, are the trustworthy half of this output -- see Section~\ref{sec:limitations}); and (ii) the AR reconstruction RMSE, overall and per axis per pole, an internals-level fidelity signal that never looks at vocabulary at all.

\paragraph{Semantic (vocabulary-based) family clustering.} None of the three probes above says whether two \emph{different} axes -- possibly discovered in different topics -- contrast the same thing. Section~\ref{sec:method-field}'s $\lvert\cos\rvert$-direction clustering answers a geometric version of this question; here we answer a meaning-based version directly. Each axis's differential-TF-IDF vocabulary is turned into one signed vector over the corpus-wide term vocabulary (positive weight for pole-B-characteristic terms, negative for pole-A, magnitude the TF-IDF mean-difference score, $\ell_2$-normalised); two axes are compared by the absolute cosine similarity of these vectors (sign-invariant, since which pole is labelled A versus B is an arbitrary bookkeeping choice -- Section~\ref{sec:method-discovery}), and clustered by the same average-linkage machinery as the geometric clustering, at merge threshold $0.3$ (chosen from the empirical distribution of pairwise similarities over this run's 543 axes: median $0.06$, 95th percentile $0.29$, 99th percentile $0.50$ -- $0.3$ is therefore a selective cut, not a threshold copied from the differently-scaled geometric case). Each family's two poles are auto-labelled by the top terms of its members' averaged vector, giving a human-readable name for free.

\paragraph{Sentiment-alignment test.} An off-the-shelf, off-domain sentiment classifier (\url{cardiffnlp/twitter-roberta-base-sentiment-latest} \citep{loureiro2022timelms}) scores every document once, reduced to a continuous scalar $s_i = p_{\text{pos}}(i) - p_{\text{neg}}(i)$. Each axis's $\texttt{sentiment\_delta} = \text{mean}(s \mid \text{pole B}) - \text{mean}(s \mid \text{pole A})$ is then correlated (Pearson and Spearman, pooled and per topic) against that axis's $\lvert\cos\rvert$ to $v^t_{CB}$ -- the direct, quantitative test of whether direction similarity to the known construct predicts an independent estimate of that construct, using a signal that never enters discovery, gating, or field assembly. Because the per-topic version runs $14$ simultaneous correlations, we treat the pooled test as primary and read the per-topic pattern under a Benjamini--Hochberg lens rather than at nominal $p<0.05$ (Section~\ref{sec:results-alignment}).

\section{Results}
\label{sec:results}

\subsection{How Many Axes Does a Topic Have?}
\label{sec:results-dim}

\begin{figure}[t]
\centering
\includegraphics[width=0.95\columnwidth]{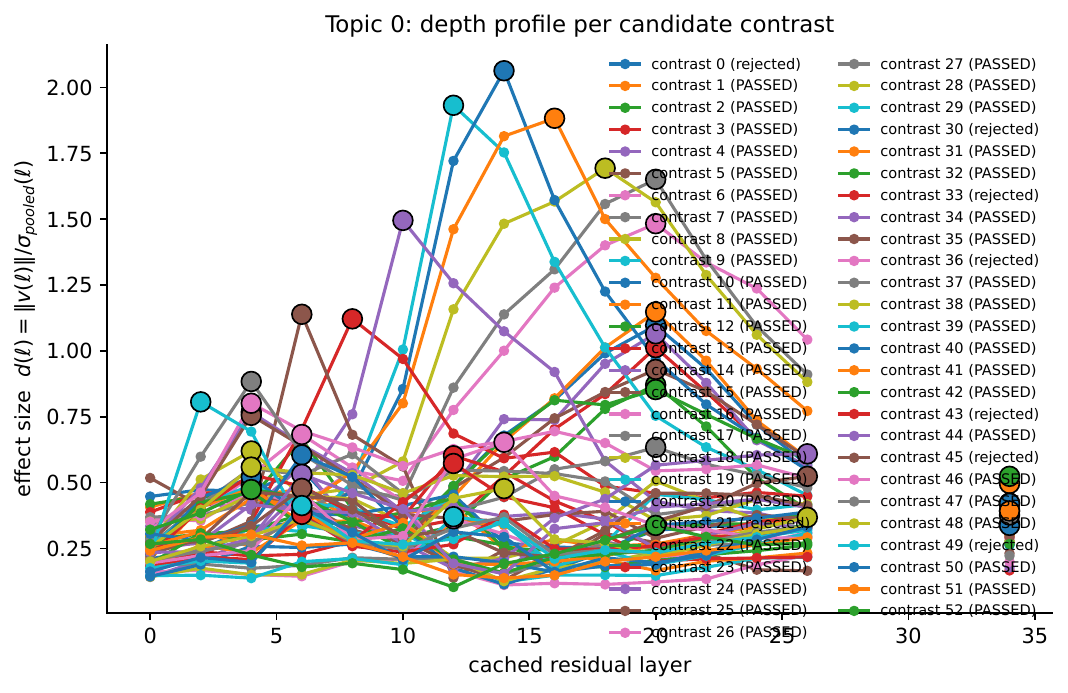}
\caption{Depth profile $d(\ell)$ for every candidate contrast discovered in one topic (Topic 0, music/CDs), across all 18 cached layers. Each curve is one candidate; most pass the stability gate (legend).}
\label{fig:depth-profiles}
\end{figure}

Per-topic rank $r_t$ ranges from $\mathbf{33}$ to $\mathbf{45}$ across the 14 topics (mean $38.8$), out of $47$--$57$ candidates proposed per topic (Table~\ref{tab:topics}). Figure~\ref{fig:depth-profiles} shows why, for one topic: dozens of candidate contrasts each trace a distinct depth profile $d(\ell)$, peaking at different layers and with different strengths, most surviving the stability gate. A single diff-in-means vector, computed at one layer, is definitionally blind to this multiplicity: it reports one line through a cloud this analysis shows to be organised along dozens of statistically independent lines.

\subsection{The Ruler Works: Blind Identification of Sentiment}
\label{sec:results-id}

In every one of the 14 topics, at least one gate-passing discovered axis matches $v^t_{CB}$ at $\lvert\cos\rvert \geq 0.706$ (mean $0.860$, range $0.706$--$0.973$; Table~\ref{tab:topics}), despite discovery never seeing the regime label. One working hypothesis, formed before running this analysis, was that dropping the 3-star middle band should make sentiment so dominant in a topic's variance that identification would always land at the \emph{first} recursion level. That hypothesis does not survive contact with the data: identification lands at recursion depth 1 in only \textbf{1 of 14} topics; the remaining 13 require dropping through 2 to 4 levels of deflation first (6 topics at depth 4, 3 at depth 3, 4 at depth 2). Even in a corpus engineered to make sentiment the easiest possible axis to find, sentiment is typically \emph{not} the direction of greatest variance in a topic's document cloud -- something else usually is, and has to be projected away first, itself indirect evidence that the excess structure reported next is not a minor perturbation on top of sentiment.

\begin{table*}[t]
\centering
\small
\begin{tabular}{@{}llrrrrrr@{}}
\toprule
\# & Topic label & Docs & Candidates $\to$ Passed & Identification $\lvert\cos\rvert$ & Found at depth & Excess axes \\
\midrule
0  & Music/CDs               & 1285 & 53 $\to$ 45 & 0.812 & 4 & 8 \\
1  & Defective/returned merch.\ & 1095 & 49 $\to$ 35 & 0.813 & 4 & 4 \\
2  & How-to/reference books    & 1077 & 48 $\to$ 36 & 0.929 & 4 & 9 \\
3  & Movies/TV (DVD)             & 1049 & 47 $\to$ 40 & 0.973 & 2 & 8 \\
4  & Home/tools/kitchen           & 831  & 57 $\to$ 42 & 0.953 & 2 & 7 \\
5  & Fiction (critical)             & 830  & 51 $\to$ 43 & 0.854 & 2 & 14 \\
6  & Consumer electronics             & 705  & 50 $\to$ 41 & 0.885 & 4 & 6 \\
7  & Religion/politics nonfiction       & 612  & 49 $\to$ 36 & 0.823 & 1 & 16 \\
8  & Fiction (enthusiastic)               & 585  & 53 $\to$ 40 & 0.706 & 3 & 5 \\
9  & Misc.\ brief positive reviews          & 557  & 56 $\to$ 43 & 0.838 & 4 & 7 \\
10 & DVD/Blu-ray quality/pricing               & 408  & 49 $\to$ 35 & 0.846 & 2 & 5 \\
11 & Food/personal-care/pet                      & 376  & 51 $\to$ 33 & 0.870 & 4 & 9 \\
12 & Children's toys/books                         & 341  & 52 $\to$ 37 & 0.945 & 3 & 6 \\
13 & Video games                                     & 220  & 54 $\to$ 37 & 0.789 & 3 & 2 \\
\midrule
\multicolumn{2}{l}{\textbf{Total / mean}}                     & \textbf{9971} & \textbf{719$\to$543} & \textbf{0.860} & -- & \textbf{106} \\
\bottomrule
\end{tabular}
\caption{Per-topic discovery yield, blind identification of the sentiment axis (cosine similarity and the recursion depth at which it was found), and excess-axis count. ``Candidates $\to$ Passed'' is candidate contrasts proposed by discovery versus contrasts surviving the stability gate. Topic labels are our own reading of sampled documents (no category metadata exists in this corpus).}
\label{tab:topics}
\end{table*}

\subsection{Excess Structure and Its Depth Signature}
\label{sec:results-depth}

\begin{figure}[t]
\centering
\includegraphics[width=0.95\columnwidth]{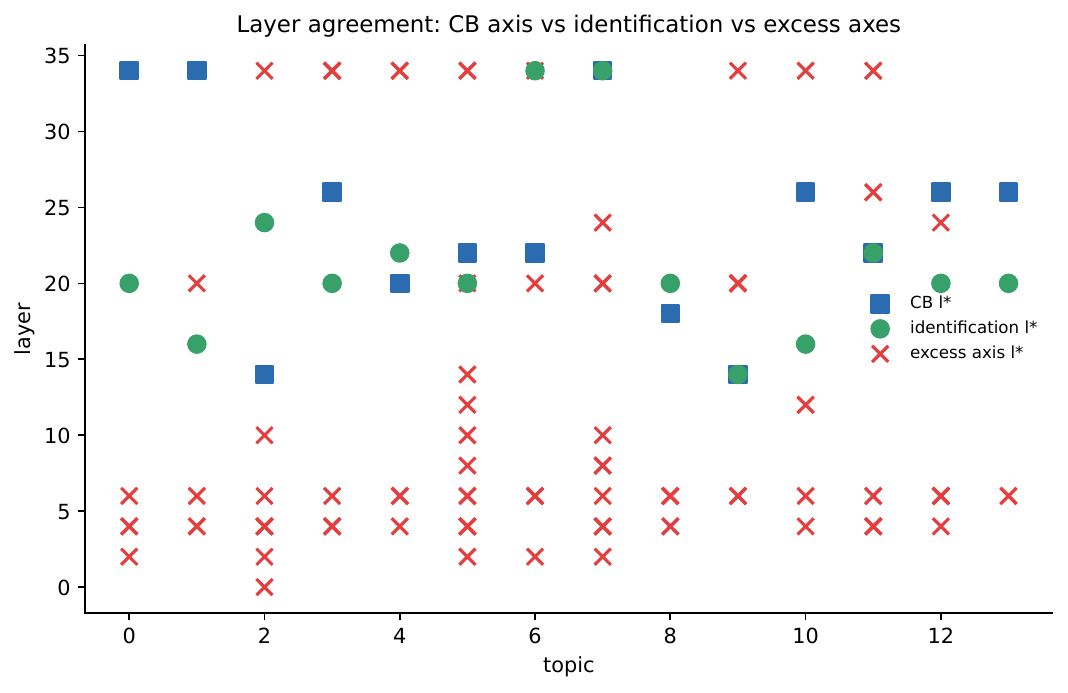}
\caption{Peak layer $\ell^\ast$ per topic for the CB axis (squares), the blindly identified matching axis (circles), and every excess axis (crosses). Sentiment consistently resolves in the middle-to-late network; excess structure concentrates early.}
\label{fig:layer-agreement}
\end{figure}

Beyond the one identification axis per topic, $\mathbf{106}$ further gate-passing axes across the 14 topics fall below the $\lvert\cos\rvert<0.3$ excess threshold (2--16 per topic, mean $7.6$; Table~\ref{tab:topics}, rightmost column) -- not near-misses on sentiment, but directions close to orthogonal to it, that still had to clear the identical split-half stability gate as the identification axis itself. Where these axes live in the network is, on its own, a striking result (Figure~\ref{fig:layer-agreement}): the CB axis and its blindly identified match cluster in the middle-to-late network (mostly layers $20$--$34$, occasionally as early as 14), while excess axes concentrate in the \emph{early-to-middle} layers (2, 4, 6, with a longer tail to 14 and a smaller number reaching the deep layers alongside sentiment). This depth separation is a directly measured property of the axes and does not depend on any interpretation of what depth means or of what the excess axes are actually about (Section~\ref{sec:results-qual} answers that separately, from text). One \emph{interpretation}, which we flag as a hypothesis we cannot yet test causally, is that if layer depth roughly indexes level of abstraction, then sentiment resolving late while other contrastive structure resolves early would be consistent with excess axes picking out more concrete, feature-level distinctions that are progressively abstracted away as the network forms an evaluative judgement specifically. We stress the causal reading is not established here -- the depth histogram is correlational, and confirming the abstraction account would require intervention (Section~\ref{sec:future}); the qualitative reading in Section~\ref{sec:results-qual} only checks the weaker, text-level claim that these early axes carry recognisably non-sentiment \emph{content}. That said, the direction of the effect is consistent with independent evidence from sparse probing that early layers carry many features in superposition while higher-level, more abstract contextual features are resolved by dedicated units in middle-to-later layers \citep{gurnee2023finding,elhage2022superposition}: an evaluative judgement is a plausible instance of exactly such a late-resolving abstraction, and a topical or register contrast (Section~\ref{sec:results-qual}) a plausible instance of an earlier-resolving, more concrete one.

\subsection{What Do the Excess Axes Mean? Three Independent Readings}
\label{sec:results-qual}

We read every excess axis three independent ways (Section~\ref{sec:method-part2}): differential TF-IDF vocabulary, a grounded language-model gloss, and an internals-grounded natural-language decoder (NLA). The three do not agree equally well, and we report exactly where each is and is not trustworthy, rather than lead with whichever looks best.

\paragraph{The language-model gloss route is weak, and we can show precisely how weak.} $84.9\%$ of the $543$ \texttt{gemma-3-4b-pt} completions reduce to a generic ``Group A/B reviews are more positive/negative,'' regardless of the axis's own TF-IDF content. Restricting to the $106$ excess axes specifically -- geometrically far from sentiment by construction -- the rate is $82.1\%$, barely lower. If this reflected real sentiment content the base model was correctly detecting, the sentiment-classifier delta $\lvert\Delta_{\text{sent}}\rvert$ (Section~\ref{sec:method-part2}) should be systematically larger for excess axes that received a sentiment-flavoured gloss than for those that did not; it is not ($0.156$ mean versus $0.103$, a small difference in the direction one would expect from a rate this close to the base rate, not a sharp discriminator). We read this as the base model defaulting to the single most probable review-language continuation almost irrespective of what it was shown, not as evidence that most excess axes are ``sentiment in another guise.'' TF-IDF vocabulary, which involves no generation at all, is the more reliable signal, and everything below leans on it first.

\begin{figure}[t]
\centering
\includegraphics[width=0.95\columnwidth]{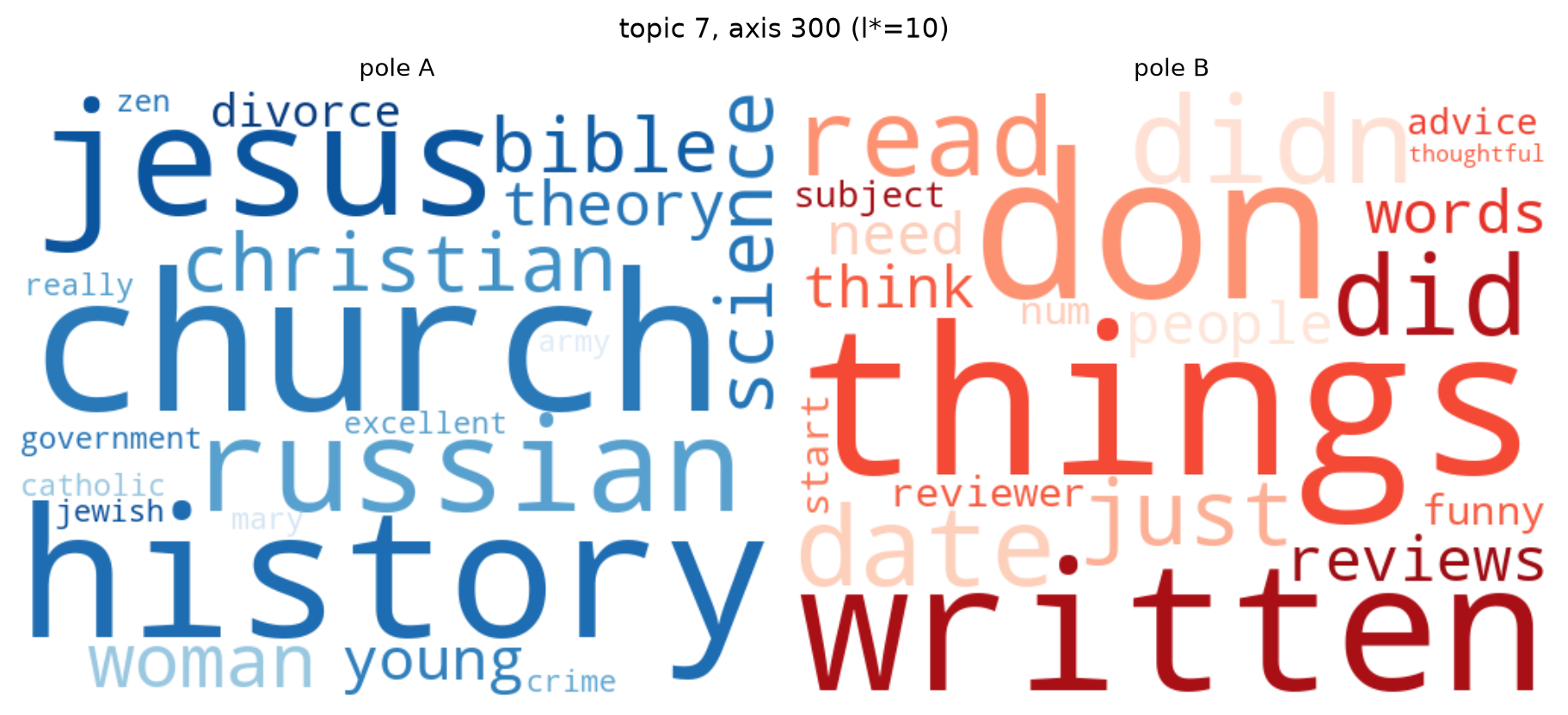}
\caption{Differential TF-IDF vocabulary for the first row of Table~\ref{tab:worked} (Topic 7, effect size $1.45$, sentiment delta $-0.010$): a content axis, not a sentiment axis.}
\label{fig:wordcloud}
\end{figure}

\paragraph{TF-IDF vocabulary and an internals-grounded decoder agree on hand-checked axes.} Table~\ref{tab:worked} gives four representative axes by their differential TF-IDF vocabulary. The first two rows (Topic 7, religion/politics/history nonfiction) have large effect size and near-zero sentiment-classifier delta ($-0.010$, $-0.009$): both poles are, on average, equally positive or negative, and the axis instead separates religious/historical subject matter (church, jesus, bible, catholic) from a more general, meta-review or humanistic register (things, written, reviewer; life, culture, government) -- a content axis, not a restatement of sentiment. The wordcloud for the first (Figure~\ref{fig:wordcloud}) shows the same split visually. The fourth row (Topic 2, how-to/reference books) is the most interesting case: its vocabulary reads as evaluative on both sides (waste/waste money vs.\ excellent/wonderful/comprehensive) and it does carry a moderate sentiment delta ($+0.306$), yet its direction is almost perfectly orthogonal to this topic's own $v^t_{CB}$ ($\lvert\cos\rvert=0.034$) -- evidence that evaluative language is not confined to a single direction even within one topic: a ``how good a reference resource is this'' judgement and the star-rating-derived sentiment direction can be geometrically distinct while both being genuinely evaluative.

The third row (Topic 6, consumer electronics) is the one we can cross-check against the NLA route directly. Its vocabulary splits computing peripherals (software, xp, windows, router, usb) from audio/photo equipment (sound, camera, headphones, lens) -- a sub-category axis, not a sentiment one (its $\lvert\cos_{CB}\rvert=0.14$ is low, and it peaks at layer 34, far from the reference layer used for the CB comparison), though it does carry a real sentiment delta ($+0.585$), plausibly because these two product sub-clusters happen to differ in average satisfaction in this corpus. Independently of any of this vocabulary analysis, the NLA encoder's learned embedding places this same axis's pole-A centroid nearest to real reviews about failing to transfer files and low-capacity RAM, and its pole-B centroid nearest to real reviews praising mini speakers and audio components -- recovering the identical computing-versus-audio split from a model trained only to reconstruct each document's own text from its activation, with no access to TF-IDF or to any other axis's vocabulary. This three-way agreement (TF-IDF, sentiment-classifier delta, and NLA's grounded nearest-neighbours, all computed independently) is the strongest single case in this paper for treating an excess axis's content as real rather than an artefact of one method's idiosyncrasies. NLA's own \emph{generated} text is a separate matter -- see Section~\ref{sec:limitations} -- but its grounding against real nearest documents does not depend on generation quality at all. Across all $543$ axes, the frozen NLA encoding supports activation reconstruction (AR) at overall $R^2=0.72$ (RMSE $252.4$, raw activation units), and the per-axis, per-pole reconstruction fidelity gap is centred near zero on average (mean $0.43$, i.e.\ no systematic bias toward either pole) but individually substantial (mean absolute gap $16.0$, roughly $7\%$ of typical RMSE) -- axis-specific, not a corpus-wide asymmetry.

\begin{table*}[t]
\centering
\small
\setlength{\tabcolsep}{4pt}
\begin{tabular}{@{}rrrrrp{4.6cm}p{4.6cm}@{}}
\toprule
Topic\# & $\ell^\ast$ & $d$ & $\lvert\cos_{CB}\rvert$ & $\Delta_{\text{sent}}$ & Top terms, pole A & Top terms, pole B \\
\midrule
7 & 10 & 1.45 & 0.29 & $-0.010$ & church, history, jesus, russian, christian, science, bible, catholic & things, don, written, didn, reviews, reviewer, thoughtful \\
7 & 20 & 0.47 & 0.33 & $-0.009$ & bible, jesus, god, bush, moore, truth, history & good, life, human, culture, government, writing, parents \\
6 & 34 & 0.45 & 0.14 & $+0.585$ & software, support, xp, windows, computer, router, usb & sound, quality, camera, price, headphones, speakers, lens \\
2 & 4 & 0.84 & 0.03 & $+0.306$ & don, waste, workout, dvd, money, waste money & excellent, information, comprehensive, wonderful, guide, helpful \\
\bottomrule
\end{tabular}
\caption{Four representative excess axes (topic \# per Table~\ref{tab:topics}: 7 religion/politics nonfiction, 6 consumer electronics, 2 how-to/reference books): peak layer $\ell^\ast$, effect size $d$ at $\ell^\ast$, absolute cosine to that topic's own CB vector, sentiment-classifier delta $\Delta_{\text{sent}}$ (near zero for the top two rows, non-trivial for the bottom two despite low $\lvert\cos_{CB}\rvert$ in all four), and the top differential TF-IDF terms per pole. Row 3 is independently corroborated by the NLA route -- see text.}
\label{tab:worked}
\end{table*}

\begin{figure}[t]
\centering
\includegraphics[width=0.95\columnwidth]{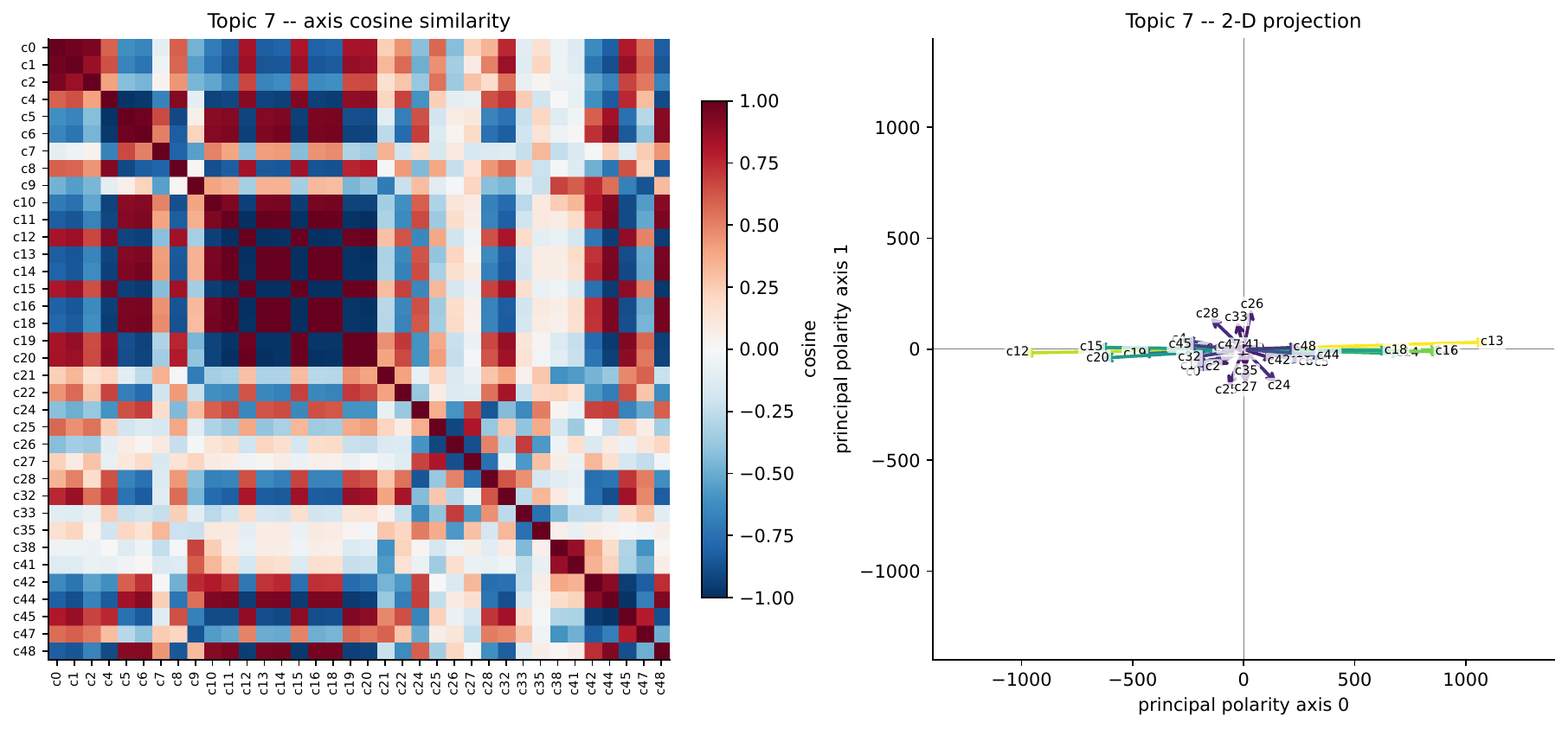}
\caption{Topic 7's own field of gate-passing axes: pairwise cosine similarity (left) and a 2-D projection onto the topic's own top two principal polarity directions (right). A large collinear block plus at least two smaller sub-clusters are visible.}
\label{fig:topic-field}
\end{figure}

Figure~\ref{fig:topic-field} shows one topic's own field of discovered axes on its own terms (Topic 7): a large block of near-collinear axes (dark red/blue -- recall sign is cosmetic, so $+1$ and $-1$ both indicate near-identical direction) corresponds to the dominant content split just discussed, alongside at least two visually distinct sub-clusters of axes -- concrete, topic-internal evidence that ``how many axes does this topic have'' is not answered by a single number, but by a structured, partly-clustered set.

\subsection{Semantic Families: What Recurs Across Topics, What Doesn't}
\label{sec:results-families}

The previous section established that individual excess axes have real, checkable content. This section asks the corpus-level question directly: do any of these specific semantic contrasts recur across topics -- candidate regularities in how people write contrastively about products in general -- or is every one of the $543$ axes a one-off, local to the topic that produced it? We answer this from vocabulary, not raw geometry, and show why that choice matters empirically, not just methodologically.

\paragraph{Clustering by shared vocabulary.} Each axis's signed differential-TF-IDF vector (Section~\ref{sec:method-part2}) is compared against every other axis's by absolute cosine similarity and clustered at threshold $0.3$, yielding \textbf{99 semantic families}. Ninety are single-axis or single-topic; \textbf{9 recur across more than one topic}, each with an auto-generated two-pole label (Figure~\ref{fig:semantic-families}; full list in Appendix~\ref{app:more-examples}). The largest, spanning three predominantly-book topics (fiction, religion/politics nonfiction, and enthusiastic fiction; $33$ axes), contrasts government/history/war/american vocabulary against book/mystery/read/excellent vocabulary -- political or historical subject matter versus narrative or genre-fiction language, a content distinction with no obvious sentiment reading. The second largest, spanning \emph{eight} topics ($24$ axes) that otherwise have little in common (music, defective merchandise, how-to books, electronics, religion/politics, food/care/pet), contrasts questions/learning/workout/didn against love/guide/baby/examples/information -- a recurring instructional-versus-generic-complaint register that is not the same family as, and does not require, sentiment. A third ($16$ axes, six topics) does read as sentiment-flavoured (bad/worst/boring versus great/love/awesome/best), and is the clearest sentiment-adjacent semantic family besides the CB-identification axes themselves -- but note that it is a \emph{different} family from any single topic's identification axis, i.e.\ this particular sentiment-flavoured vocabulary cluster is itself a piece of excess structure, not a restatement of the ruler.

\begin{figure}[t]
\centering
\includegraphics[width=0.95\columnwidth]{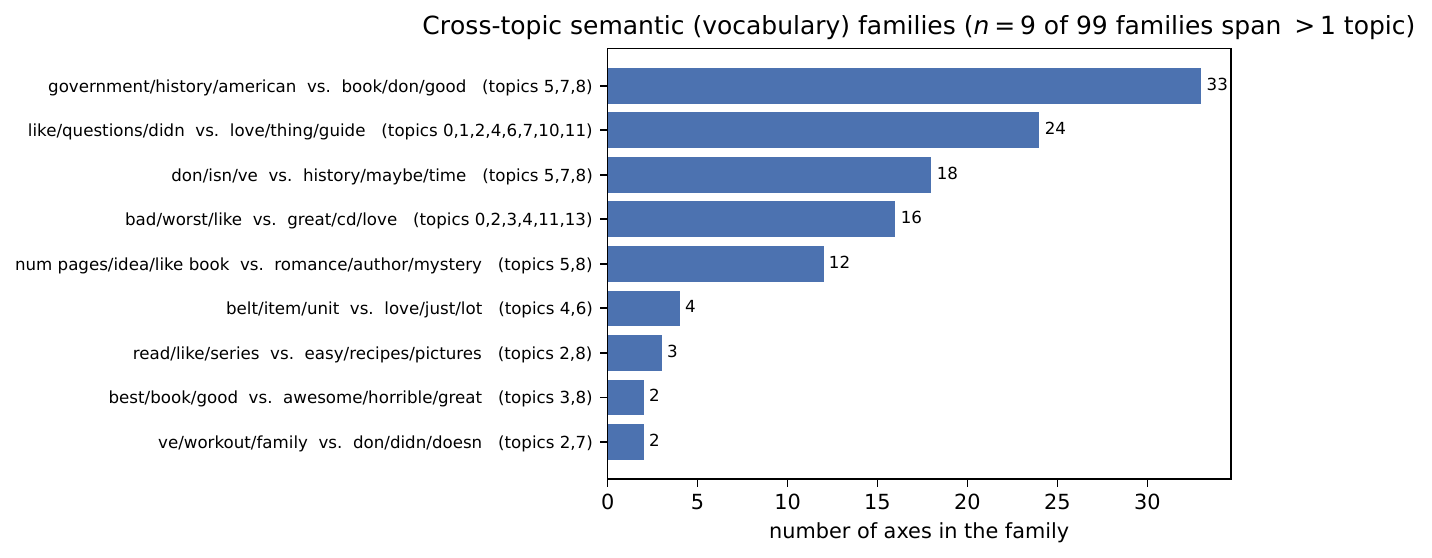}
\caption{The 9 semantic (vocabulary-based) families spanning more than one topic, out of 99 total, with their auto-generated two-pole labels (top 3 terms per pole) and topic membership (topic numbers per Table~\ref{tab:topics}).}
\label{fig:semantic-families}
\end{figure}

\paragraph{This is not the same partition raw direction gives.} Clustering the same $543$ axes by $\lvert\cos\rvert$ similarity of their raw activation-space direction (not vocabulary), pooled at the single fixed reference layer $\ell_{\text{ref}}=16$, gives a superficially similar-sounding picture -- \textbf{90 geometric families}, one spanning all 14 topics -- but the two clusterings are close to independent: agreement between them is chance-level (adjusted Rand index $=0.005$). Pooling the same $543$ axes' raw vectors also gives a participation ratio of only $\mathrm{PR}=1.18$, the first principal direction alone explaining $91.9\%$ of pooled variance, even though each topic individually supports $33$--$45$ orthogonal axes (Section~\ref{sec:results-dim}) -- i.e.\ at one fixed layer, nearly all $543$ axes' raw directions look almost the same, which cannot be literally true of their content given Section~\ref{sec:results-qual}'s worked examples. We read this as an artefact of the comparison, not a deeper fact about the field: every axis is evaluated at $\ell_{\text{ref}}=16$ regardless of its own peak layer $\ell^\ast$ (Table~\ref{tab:worked}'s four axes alone span $\ell^\ast \in \{4,10,20,34\}$), so axes that are geometrically distinctive \emph{at their own best layer} can look nearly parallel at a layer chosen for something else (sentiment separability) entirely. Raw-direction clustering and pooled-PR dimensionality are worth reporting because they are the natural extension of the diff-in-means machinery this paper's ruler itself uses, but Section~\ref{sec:results-qual} and the semantic clustering above are, on this evidence, the more trustworthy answer to ``what recurs.'' We return to what the near-zero agreement itself implies in Section~\ref{sec:discussion}.

\subsection{Does Direction Alone Predict Meaning?}
\label{sec:results-alignment}

\begin{figure}[t]
\centering
\includegraphics[width=0.95\columnwidth]{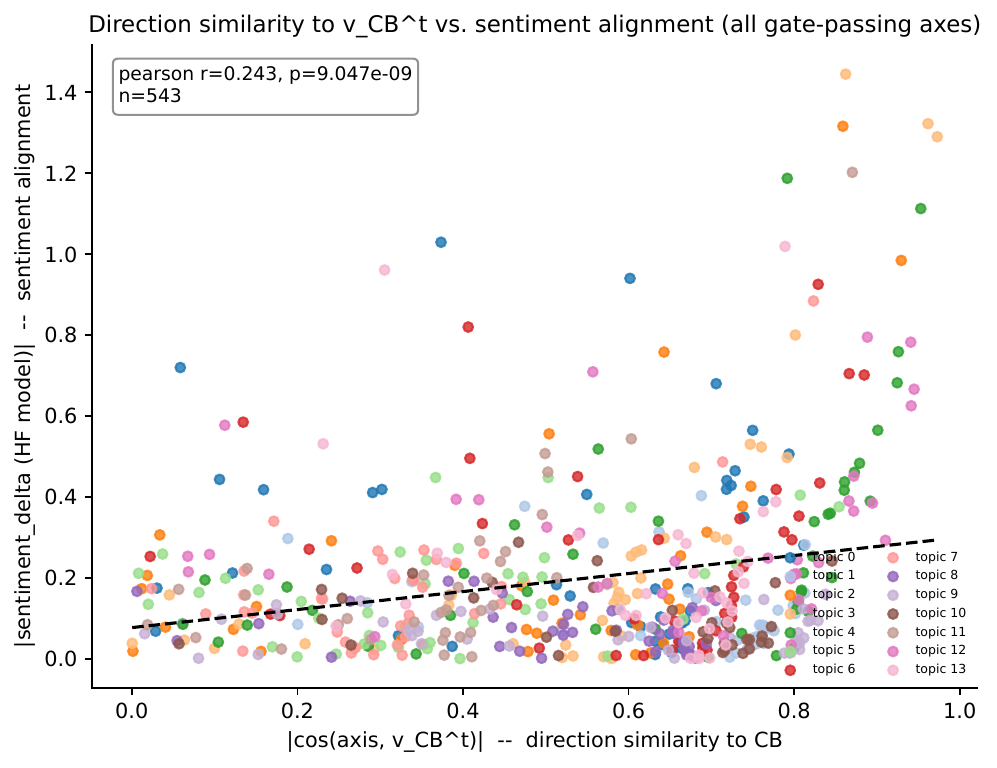}
\caption{Direction similarity to $v^t_{CB}$ versus independent sentiment-classifier alignment, all 543 gate-passing axes, coloured by topic. Positive and highly significant pooled, but explaining under 6\% of variance.}
\label{fig:alignment}
\end{figure}

The sharpest test of TOPol's central claim is quantitative, not anecdotal: if an axis's direction similarity to the known construct ($\lvert\cos_{CB}\rvert$) predicted an \emph{independent} measure of that construct, direction alone would be sufficient to read an axis's meaning off its geometry, and the qualitative diversity in Section~\ref{sec:results-qual} would be a curiosity rather than evidence of real additional dimensions. Pooled across all $543$ axes, $\lvert\cos_{CB}\rvert$ correlates with the sentiment classifier's $\lvert\Delta_{\text{sent}}\rvert$ at $r=0.243$ (Pearson, $p=9.05\times10^{-9}$; Spearman $\rho=0.155$, $p=2.90\times10^{-4}$) -- positive and highly significant, but explaining under $6\%$ of variance; Figure~\ref{fig:alignment} shows the scatter directly, including a substantial population of low-$\lvert\cos_{CB}\rvert$ axes with large sentiment deltas and high-$\lvert\cos_{CB}\rvert$ axes with near-zero deltas. The pooled test also hides considerable heterogeneity (Table~\ref{tab:peralignment}): six of 14 topics show a strong positive correlation ($r=0.38$ to $0.69$), one a \emph{negative} correlation, and the remaining seven no relationship. Because this is $14$ simultaneous tests, we read it under Benjamini--Hochberg control rather than at nominal $p<0.05$; the three strongest positive topics ($r\ge0.65$, $p\le2\times10^{-5}$) survive any reasonable correction and the pooled result does not depend on the per-topic significance pattern at all -- the heterogeneity itself, not the count of ``significant'' topics, is the finding. Direction similarity to a known construct is therefore real, measurable, and statistically robust in aggregate, but far too weak and too topic-dependent to serve as a stand-alone proxy for what an axis means -- supporting, quantitatively, the qualitative pattern in Section~\ref{sec:results-qual} that many axes carry non-redundant information sentiment geometry alone does not capture. (The rating-correlation half of this test would use the continuous star rating directly rather than an off-the-shelf classifier; the corpus preparation used here exposes only a precomputed binary label, so that half could not be computed, $n=0$, and the classifier-based test above is the one we report throughout.)

\begin{table}[t]
\centering
\small
\begin{tabular}{@{}lrrr@{}}
\toprule
Topic & $n$ & $r$ & $p$ \\
\midrule
3 Movies/TV        & 40 & $\mathbf{0.680}$ & $1.4\times10^{-6}$ \\
11 Food/care/pet     & 33 & $\mathbf{0.686}$ & $1.0\times10^{-5}$ \\
7 Religion/politics    & 36 & $\mathbf{0.647}$ & $2.0\times10^{-5}$ \\
4 Home/tools              & 42 & $\mathbf{0.473}$ & $1.6\times10^{-3}$ \\
2 How-to/reference           & 36 & $\mathbf{0.434}$ & $8.2\times10^{-3}$ \\
5 Fiction (critical)            & 43 & $\mathbf{0.383}$ & $1.1\times10^{-2}$ \\
8 Fiction (enthusiastic)          & 40 & $-0.355$ & $2.5\times10^{-2}$ \\
1 Defective/returns                & 35 & $-0.218$ & 0.21 \\
10 DVD/Blu-ray quality               & 35 & $-0.225$ & 0.19 \\
13 Video games                        & 37 & $-0.182$ & 0.28 \\
12 Children's toys                     & 37 & 0.187 & 0.27 \\
9 Misc.\ brief positive                  & 43 & $-0.120$ & 0.44 \\
0 Music/CDs                                & 45 & $-0.101$ & 0.51 \\
6 Consumer electronics                      & 41 & 0.066 & 0.68 \\
\bottomrule
\end{tabular}
\caption{Per-topic Pearson correlation between $\lvert\cos_{CB}\rvert$ and $\lvert\Delta_{\text{sent}}\rvert$. Bold: individually significant at $p<0.05$ (six positive, one negative, seven not significant).}
\label{tab:peralignment}
\end{table}

\section{Discussion}
\label{sec:discussion}

\subsection{Polarity as a Property of Geometry, Not of a Document}

The results above support the paper's central reframing directly. \textbf{Existence}: every topic supports $33$--$45$ mutually orthogonal, independently reproducible bipolar splits (Section~\ref{sec:results-dim}), not one. \textbf{Calibration}: a blind procedure recovers the one axis we know the answer for, in every topic, without being told what to look for (Section~\ref{sec:results-id}) -- which is what licenses trusting what else it finds. \textbf{Excess is real, not noise, and not always evaluative}: $106$ axes pass the identical stability gate as sentiment while sitting far from it in direction (Section~\ref{sec:results-depth}); on hand-checked axes, three independent readings -- TF-IDF vocabulary, sentiment-classifier delta, and an internals-grounded decoder's nearest-neighbour grounding -- agree on content that has nothing to do with evaluation (computing versus audio-equipment reviews; religious/historical versus general-nonfiction subject matter), while a fourth reading (a non-instruction-tuned language model's free gloss) is shown quantitatively to default to a generic sentiment framing almost independently of what it is shown, and should not be read as evidence that these axes are secretly about sentiment after all (Section~\ref{sec:results-qual}). \textbf{Layer is a semantic variable}: sentiment consistently resolves in middle-to-late layers while excess structure concentrates earlier (Figure~\ref{fig:layer-agreement}), a pattern invisible to any method that fixes one analysis layer in advance. \textbf{Recurring versus topic-local, measured from meaning rather than geometry}: clustering axes by shared distinctive vocabulary -- not raw direction -- finds $9$ semantic families (of $99$) that recur across topics, each with an auto-generated, human-readable label, from a content contrast (political/historical versus narrative vocabulary, three topics, $33$ axes) to a register contrast (instructional versus generic-complaint language, eight topics, $24$ axes) to one genuinely sentiment-flavoured family that is itself distinct from any topic's own identification axis (Section~\ref{sec:results-families}). This vocabulary-based clustering agrees with the geometric (raw-direction, single-fixed-layer) clustering at only chance level (ARI $=0.005$) -- which we read not as either method failing, but as evidence that ``points the same way in the residual stream'' and ``contrasts the same vocabulary'' are close to independent properties of a discovered axis, so a geometric family structure and a semantic one answer genuinely different questions and neither should be read as a proxy for the other. Whether any of the $9$ recurring semantic families is a genuine \emph{cross-corpus} universal, comparable in status to sentiment, is a question this single-corpus study poses but cannot answer; it is the natural target of the discovery-condition sequel.

\subsection{Limitations and Scope}
\label{sec:limitations}

We separate what this calibration study \emph{deliberately brackets} (and the companion discovery study will address) from what would strengthen the present claims directly. Each item below is stated with the interpretation it does and does not license.

\paragraph{The language-model gloss route needs the quantitative check we gave it, not intuition.} $84.9\%$ of the $543$ \texttt{gemma-3-4b-pt} completions reduce to ``more positive/negative,'' and we do not simply disclose this rate: Section~\ref{sec:results-qual} shows the rate is nearly as high ($82.1\%$) among axes that are geometrically excess by construction, and that a generic gloss does not track lower independent sentiment content than a non-generic one. This is a real limitation of a small, non-instruction-tuned base model asked to gloss freely, not evidence against the axes' distinctiveness, which TF-IDF and NLA establish independently. \texttt{gemma-3-4b-it} (Section~\ref{sec:method-part2}) is configured as a direct comparison but not run for the results reported here.

\paragraph{NLA's free-generated text is frequently degenerate; its grounding is not.} The GRU decoder's own generated completions -- especially at the interpolated/extrapolated points of a pole-to-pole sweep, never seen in training -- are often repeated boilerplate or dominated by out-of-vocabulary tokens, a known failure mode of a small decoder trained on truncated per-axis text. We do not rely on this generated text as evidence anywhere in Section~\ref{sec:results-qual}; the claims we do make rest on the encoder's learned embedding space (pole centroids and their nearest REAL documents) and on the separately-trained AR head's reconstruction fidelity, neither of which depends on decoding quality.

\paragraph{The semantic clustering is a bag-of-words method.} Two axes that contrast genuinely similar content in different words (a synonym pair, or a paraphrase) will not be merged, so the $9$ recurring families in Section~\ref{sec:results-families} are a lower bound on how much semantic recurrence exists, not a complete inventory; the merge threshold ($0.3$) and the small custom stop-term list (corpus placeholder tokens only) are both reported choices, not tuned against an outcome, but neither has been stress-tested against alternatives.

\paragraph{One model, one corpus, one condition.} Every number here comes from one model (Gemma 3 4B, pretrained), one corpus (Amazon reviews), and the calibration condition specifically -- a corpus where the primary axis is known in advance, by design. The complementary discovery condition, where the primary axis is \emph{not} known in advance, is future work (Section~\ref{sec:future}).

\paragraph{Correlational, not causal, interpretation.} Every reading in Section~\ref{sec:results-qual} is drawn from which documents fall at an axis's extremes, what vocabulary separates them, or what a model trained to reconstruct them encodes -- informative, and cross-checked three independent ways, but none of the three decomposes what network features compose a direction, and none tests what generation does under intervention along it.

\section{Conclusion and Future Work}
\label{sec:future}

TOPol recasts polarity discovery as an inverse problem: instead of mapping a label to a direction, it maps a topic to the full set of directions its documents support, and uses a trusted label as a ruler rather than a target. This paper is deliberately the \emph{calibration} half of that program: on a corpus engineered to be maximally favourable to the standard sentiment-is-polarity equation, the ruler confirms itself (every topic's sentiment axis is blindly recoverable) while simultaneously proving itself insufficient -- $106$ further axes pass the same reproducibility bar, concentrate at different network depths than sentiment does, are only weakly and heterogeneously predicted by direction similarity to the sentiment vector, and include, on independent cross-checked evidence, genuine non-evaluative content (subject matter, product sub-category) as well as evaluative structure that is itself distinct from the sentiment ruler. Clustering by what these axes actually contrast, rather than by raw direction pooled at one layer, finds a small number of specific semantic regularities that recur across topics and a large majority that do not -- the beginning of an empirical typology, not an assertion of one.

Three classes of next step follow. \textbf{Strengthening the present claims.} A direct selectivity baseline for the stability gate -- how many random, PCA, or label-shuffled directions clear it -- would convert the high-dimensional-cosine argument of Section~\ref{sec:method-gate} into a measured false-pass rate; a permutation null for the pooled participation ratio would do the same for the pooled-geometry dimensionality claim; replicating on a second model (e.g.\ a different-size checkpoint of the same family) would establish that the field is not a Gemma-3-4B artifact; and completing the \texttt{gemma-3-4b-it} comparison already configured (Section~\ref{sec:method-part2}) would test directly whether an instruction-tuned model resolves more of the $84.9\%$ generic base-model completions, or merely confabulates differently. \textbf{Deepening interpretation.} Sparse-autoencoder decomposition of the discovered axes against a released dictionary for the same model \citep{bricken2023monosemanticity,templeton2024scaling,cunningham2023sparse} would move from correlational to compositional interpretation of what each axis is built from; causal steering along each axis, adding and subtracting it during generation \citep{turner2023steering,rimsky2024steering}, would test whether the depth-abstraction hypothesis of Section~\ref{sec:results-depth} holds under intervention rather than merely in correlation; and stress-testing the semantic-family clustering itself (alternative merge thresholds, embedding-based rather than bag-of-words similarity, so that synonymous but lexically distinct axes can be recognised as related) would tighten Section~\ref{sec:results-families}'s typology. \textbf{The discovery condition.} Finally, running the identical pipeline where the method is ultimately aimed -- a corpus such as central-bank communications around a regime break, where the primary axis of contrast is \emph{not} specified in advance and the typology in Section~\ref{sec:results-families} is the finding rather than a validation check -- is the payoff the calibration here is meant to license.

\section*{Ethics and Data Statement}
This work analyses publicly-authored product reviews and does not collect, publish, or attempt to re-identify any reviewer; excerpts reproduced in this paper and its appendix are used for scientific illustration under the same terms as the source corpus. All models used are openly released and run locally; no proprietary API or third-party data-labelling service was used at any stage.

\bibliographystyle{topol}
\bibliography{topol}

\clearpage
\appendix
\section{Additional Configuration Detail}
\label{app:config}

Full run configuration is serialised with the run (\texttt{config.json}) and summarised here for reference. Discovery: $\texttt{layer\_stride}=2$, $\texttt{max\_depth}=4$, $\texttt{min\_cluster\_size}=30$, $\texttt{ari\_merge}=0.7$, $\texttt{stability\_threshold}=0.6$, $\texttt{excess\_cos\_threshold}=0.3$, $\texttt{ref\_layer}=16$ (the layer chosen by probe separability). Segmentation: EmbeddingGemma-300m, UMAP ($k=15$, cosine, 5 components), Leiden (resolution $1.0$), boundary-independent (fit on regime A, nearest-centroid assign regime B). TF-IDF: unigrams+bigrams, English stop-words, $\texttt{min\_df}=2$, $3{,}000$ max features, top $20$ terms per pole. Language-model glosses: $4$ excerpts per pole (max $40$ words each), $\texttt{max\_new\_tokens}=50$, batch size $4$--$8$. NLA/AR: latent dimension $256$, encoder/decoder hidden dimension $512$, $150$ training epochs (Adam, learning rate $10^{-3}$), AR head under $5$-fold cross-validation, $60$ epochs per fold. Semantic family clustering: signed, $\ell_2$-normalised TF-IDF-difference term vectors, average-linkage clustering on $1-\lvert\cos\rvert$ distance, merge threshold $0.3$. Sentiment classifier: \url{cardiffnlp/twitter-roberta-base-sentiment-latest}, batch size $32$, max sequence length $128$. All stochastic steps (topic segmentation, 2-means initialisation, half-splits, document sampling for prompts) use a single fixed corpus-level random seed ($42$), with per-step deterministic offsets (e.g.\ $\texttt{random\_state}{+}1$/$+2$ for the two gate halves) so re-running the pipeline against the same cached activations reproduces every reported number exactly; this is verified live at run time by an explicit bit-identical check on the CB field estimator. Software: Python, PyTorch ($\geq2.1$), Hugging Face \texttt{transformers} ($\geq4.42$), \texttt{scikit-learn} ($\geq1.3$), \texttt{umap-learn} ($\geq0.5.4$), \texttt{scikit-network} ($\geq0.31$) for Leiden, and \texttt{scipy}/\texttt{pandas} for the statistical tests, pinned as minimum versions in the run's environment specification. Hardware: one CUDA GPU sufficient to hold \texttt{gemma-3-4b-pt} in bfloat16 plus the sentiment classifier; the exact GPU model, memory, and driver/CUDA version were not captured in this run's manifest and are not reported here as a result.

\section{Additional Worked Examples}
\label{app:more-examples}

Table~\ref{tab:pooled-geo} reports the geometric ($\lvert\cos\rvert$-direction) cross-topic family sizes underlying Section~\ref{sec:results-families}'s pooled-geometry discussion in full; Table~\ref{tab:sem-families-full} reports all $9$ cross-topic semantic (vocabulary-based) families, extending Figure~\ref{fig:semantic-families} with their full auto-generated labels. The two tables' family memberships do not correspond row-for-row (Section~\ref{sec:results-families} reports their agreement is chance-level). Full per-axis tables (differential TF-IDF terms, language-model glosses, NLA pole prototypes/nearest-neighbours, and the complete document text behind every pole of every one of the 543 axes) are produced by this pipeline as CSV artifacts (\texttt{topol\_pv\_tfidf\_summary}, \texttt{topol\_pv\_llm\_explanations}, \texttt{topol\_nla\_pole\_prototypes}, \texttt{topol\_sem\_families}, \texttt{topol\_pv\_full\_texts}) and released alongside the code (Appendix~\ref{app:code}) rather than reproduced in full here.

\begin{table}[h]
\centering
\small
\begin{tabular}{@{}rrl@{}}
\toprule
Family & \# topics & Topics \\
\midrule
84 (universal) & 14 & all \\
80 & 8 & 0,2,3,5,8,9,10,12 \\
78 & 6 & 0,3,4,5,11,13 \\
64 & 5 & 0,1,6,8,12 \\
63 & 4 & 0,2,5,7 \\
12 & 3 & 0,3,8 \\
\midrule
84 other families & 1 each & (topic-local) \\
\bottomrule
\end{tabular}
\caption{The six cross-topic \emph{geometric} ($\lvert\cos\rvert$-direction) families; the remaining 84 of 90 such families are single-topic. Pooled at the single fixed reference layer $\ell_{\text{ref}}=16$ -- see Section~\ref{sec:results-families} for why this is a limited view of ``what recurs.''}
\label{tab:pooled-geo}
\end{table}

\begin{table}[h]
\centering
\small
\setlength{\tabcolsep}{3pt}
\begin{tabular}{@{}rrlp{3.4cm}@{}}
\toprule
Fam. & Axes & Topics & Pole A terms / Pole B terms \\
\midrule
73 & 33 & 5,7,8 & government, history, american / book, don, good \\
30 & 24 & 0,1,2,4,6,7,10,11 & like, questions, didn / love, thing, guide \\
72 & 18 & 5,7,8 & don, isn, ve / history, maybe, time \\
42 & 16 & 0,2,3,4,11,13 & bad, worst, like / great, cd, love \\
60 & 12 & 5,8 & num pages, idea, like book / romance, author, mystery \\
31 & 4 & 4,6 & belt, item, unit / love, just, lot \\
41 & 3 & 2,8 & read, like, series / easy, recipes, pictures \\
33 & 2 & 2,7 & ve, workout, family / don, didn, doesn \\
77 & 2 & 3,8 & best, book, good / awesome, horrible, great \\
\bottomrule
\end{tabular}
\caption{All 9 cross-topic \emph{semantic} (vocabulary-based) families, extending Figure~\ref{fig:semantic-families}: family id, axis count, topics spanned (numbering per Table~\ref{tab:topics}), and each pole's top 3 auto-generated terms.}
\label{tab:sem-families-full}
\end{table}

\section{Code and Data Availability}
\label{app:code}
\enlargethispage{\baselineskip}
All code (topic segmentation, discovery, gating, field assembly, differential-vocabulary and alignment analysis, and every figure and table in this paper) is publicly available under the MIT license at \url{https://github.com/gomez-L/TOPol}, together with the data needed to reproduce the calibration case reported here and the exact run manifest, so that every reported number can be regenerated without re-running the GPU-bound activation extraction step. The underlying Amazon review text and star-rating labels are drawn from a pre-existing, publicly-sourced product-review corpus and are not themselves a contribution of this paper.

\end{document}